\documentclass{article}
\usepackage{xurl}
\usepackage[english]{babel}

\usepackage[a4paper]{geometry}

\usepackage{float}
\usepackage{amsmath}
\usepackage[ruled,vlined,algo2e]{algorithm2e}
\usepackage{amsfonts} 
\usepackage{graphicx}
\usepackage[colorlinks=true, allcolors=blue,pdfusetitle=true]{hyperref}

\usepackage[nottoc]{tocbibind}
\usepackage{bibentry}
\nobibliography*
\usepackage{authblk}

\usepackage{enumitem}
\usepackage{xcolor}

\title{Monte-Carlo Tree Search with Neural Network Guidance \\ for Lane-Free Autonomous Driving}

\author{Ioannis Peridis}
\author{Dimitrios Troullinos}
\author{Georgios Chalkiadakis}
\author{Pantelis Giankoulidis}
\author{Ioannis Papamichail}
\author{Markos Papageorgiou}
\affil{Technical University of Crete}
\affil{\{iperidis, dtroullinos, gchalkiadakis, pgiankoulidis, ipapamichail, mpapageorgiou\}@tuc.gr}
\date{}

\begin{document}
	
\maketitle
\begin{abstract}
Lane-free traffic environments allow vehicles to better harness the lateral capacity of the road without being restricted to lane-keeping, thereby increasing the traffic flow rates.
As such, we have a distinct and more challenging setting for autonomous driving.
In this work, we consider a Monte-Carlo Tree Search (MCTS) planning approach for single-agent autonomous driving in lane-free traffic, where the associated Markov Decision Process we formulate is influenced from existing approaches tied to reinforcement learning frameworks. 
In addition, MCTS is equipped with a pre-trained neural network (NN) that guides the selection phase. 
This procedure incorporates the predictive capabilities of NNs for a more informed tree search process under computational constraints. 
In our experimental evaluation, we consider metrics that address both safety (through collision rates) and efficacy (through measured speed). 
Then, we examine: (a) the influence of isotropic state information for vehicles in a lane-free environment, resulting in nudging behaviour---vehicles' policy reacts due to the presence of faster tailing ones, 
(b) the acceleration of performance for the NN-guided variant of MCTS, 
and (c) the trade-off between computational resources and solution quality.
\end{abstract}

\section{Introduction}

The technological progress in autonomous vehicles, introducing as it does fast and reliable observational and connectivity capabilities, gives rise to research endeavours that envision forthcoming traffic environments with full autonomy.
Among these, {\em lane-free traffic}~\cite{Papageorgiou2021} is a recently introduced paradigm that lifts the requirement for lane-keeping behaviour.
In this manner, vehicles can operate freely 
within the road boundaries, and better harness road capacity~\cite{Papageorgiou2021}.

One of the investigated vehicle movement strategies puts forward a Markov Decision Process (MDP)~\cite{Karalakou2023} for a single-agent formulation attached to (deep) reinforcement learning algorithms.
An important aspect in lane-free autonomous driving is the notion of {\em nudging}, meaning that vehicles can observe both front and back-located vehicles, and react accordingly (e.g., a vehicle making space to assist a faster tailing vehicle).
This is implicitly incorporated in the MDP of~\cite{Karalakou2023}, but its effect is not actually examined; whereas it is empirically found to greatly improve upon the resulting policy for other methods~\cite{Papageorgiou2021}.
To the best of our knowledge, no work to date tackles planning-based policies in lane-free traffic.

Against this background, in this paper we propose a refined MDP that is better suited to planning approaches, 
and put forward a systematic investigation of the potential created by using the celebrated Monte-Carlo Tree Search (MCTS) planning method~\cite{Browne2012MCTSSurvey} for {\em lane-free autonomous driving}.
Moreover, we provide the framework for the integration of neural networks (NNs) in MCTS planning in this domain.
To this end, we incorporate prior knowledge in our MDP with NNs and properly calibrate them with respect to MCTS-generated data,
building on well-established lines of work~\cite{Kemmerling2023} that incorporate NNs with prior knowledge (e.g., an expert policy) in the design
of MCTS planning approaches.
This integration can reduce the necessity for extensive computational efforts, i.e., achieve the same solution quality  much faster. 
To this end, we employ an NN-Guided MCTS method~\cite{Kemmerling2023}, with a leveraged NN trained  
via offline self-play simulations in lane-free environments, utilizing data produced by our MCTS algorithm.
With this, we can effectively determine the acceleration benefit of NNs in lane-free traffic, and 
discuss the computational trade-offs when NNs are involved.

We provide a systematic experimental evaluation of the trade-offs between computational effort and solution quality for lane-free autonomous driving, with metrics that assess safety and speed preferences in a challenging environment.
Our results show that isotropic state information has a great impact on the decision-making, specifically resulting in nudging behaviour for vehicles, and 
high performance in challenging, high density traffic scenarios.
Additionally, the NN-guided MCTS, significantly speeds up the tree search process in lane-free traffic, 
allowing the algorithm to exploit prior knowledge of the NN
and return more informed results in fewer iterations.

\section{Background and Related Work}\label{sec:back_rw}
In this section, we introduce background information and discuss related work.
\subsection{MDPs and MCTS with Neural Guidance}
A Markov Decision Process (MDP)~\cite{Bellman1957Markovian} models decision-making in stochastic sequential environments using a $5$-tuple $(S, A, P, R, \gamma)$, where \(S\) is the state space, \(A\) is the action space, \(P\) is the transition probability, \(R\) is the reward function, and \(\gamma\) is the discount factor.

Monte-Carlo Tree Search (MCTS)~\cite{Browne2012MCTSSurvey,Swiechowski2022MCTSReview} is a solution method for MDPs based on planning, and it operates by simulating action paths to estimate long-term rewards, balancing exploration and exploitation.
A search tree is built, using Monte-Carlo simulations to improve predictions. 
The tree expands iteratively through four phases---selection, expansion, simulation and backpropagation---emphasizing computational effort on promising areas of the decision space.
For the selection phase, MCTS (typically) uses the Upper Confidence Bound for Trees (UCT)~\cite{Kocsis2006BanditMCPlanning} formula, where a given a state \(s\) and a candidate action \(a\) at a tree node, the score is calculated as:
    $\text{UCT}(s, a) = Q(s, a) + C \sqrt{\frac{\ln N(s)}{N(s, a)}}$
, where $Q(s, a)$ is the estimated expected average reward that is also updated iteratively according to the simulated outcomes, $N(s)$ is the state visit count, $N(s, a)$ is the action count, and $C$ balances exploitation (optimize over the current estimate $Q(s,a)$) and exploration (expand the tree uniformly).

The embedding of deep learning into MCTS has been previously demonstrated in studies on Computer Go~\cite{Kemmerling2023} and Hex~\cite{GaoHaywardMuller2017}. 
These studies used neural networks (NNs)~\cite{GoodfellowBengioCourville2016DeepLearning} to improve move prediction and evaluation by incorporating {\em neural guidance} into the selection phase of MCTS.
NNs are typically pre-trained in a supervised manner from available datasets of expert players~\cite{Kemmerling2023}, which then can guide the planning tree.
This training produces a model that predicts action probabilities from given states. 
Once trained, the NN guides MCTS in real-time, combining the prior knowledge of the NN with the dynamic search exploration of the MCTS.

\begin{figure}[t]
    \centering
    \includegraphics[width=0.7\textwidth, keepaspectratio]{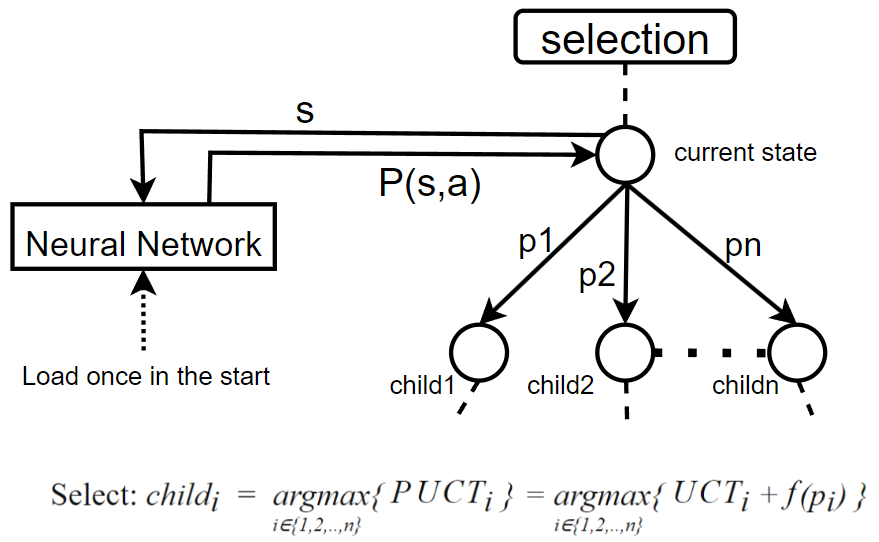} 
    \caption{Illustration of NN-Guided selection, where 
    the $child_i$ with the highest PUCT value is 
    selected.}
    \label{fig:nn_guidance}
\end{figure}

This is accomplished with 
PUCT~\cite{Kemmerling2023, GaoHaywardMuller2017}, an extension of UCT formula where selection can
leverage the NN's probability distribution \( P(s, a) \) to evaluate action success, transforming the basic UCT search into a more sophisticated, {\em NN-guided} process.
The PUCT formula, particularly variant $4$ from~\cite{Kemmerling2023}, for a state \(s\) and action \(a\) is: 
\begin{equation}\label{eq:2.19}
    \text{score}(s,a) = Q(s,a) + C_b \cdot \sqrt{\frac{\ln N(s)}{N(s, a)}} + C_{pb} \cdot \frac{P(s, a)}{N(s,a) + 1}
\end{equation}

\noindent where  \( P(s, a) \) is the NN-predicted  probability of action  \( a \) in state  \( s \), introducing neural guidance into MCTS.
Parameters \( C_b \) and \( C_{pb} \) balance exploitation, exploration, and neural guidance.
Figure~\ref{fig:nn_guidance} provides a visual illustration of Equation~\ref{eq:2.19}.

Integrating  \( P(s, a) \) into MCTS directs the search towards more promising actions. This leads to enhanced exploration efficiency, accelerates convergence to best strategies by quickly identifying promising policies, and requires less computational time. 
NN-Guided MCTS also improves decision quality under computational constraints, achieving high-quality decisions with fewer iterations or better solutions with the same number of iterations compared to plain MCTS.
While NN-guided MCTS offers advantages, it also presents challenges, primarily due to the computational demands of NNs prediction.
As mentioned in~\cite{GaoHaywardMuller2017}, this can cause an overhead that slows down the whole process, especially when dealing with deep architectures in real-time decision-making environments. 

\subsection{Related Work}
MCTS methods, known for their incremental search capabilities, are applied in autonomous driving for simulating vehicle behavior considering human intentions~\cite{Ha2020}, using KB-Trees~\cite{Chen2020}, and employing probabilistic models to predict driver actions~\cite{Lei2021}. 
They are also used for maneuver prediction based on image inputs~\cite{Volpi2017}.

Lane-free traffic considers environments where vehicle movement strategies control vehicles without being restricted to lane-keeping.
This transition to a more unstructured environment introduces complexities for motion planning, and requires innovative methods.
The principal work~\cite{Papageorgiou2021} that introduces the concept of lane-free traffic environments suggested heuristic rules to simulate ``forces'' for vehicle maneuvering, while~\cite{Karafyllis2021} developed a cruise control system using Control Theory. 
Optimal control strategies using model predictive control were presented in~\cite{Yanumula2022}, and~\cite{Troullinos2021} used the max-plus algorithm for multiagent decision making under a collaborative framework.
Additionally,~\cite{Karalakou2023} developed an MDP framework for lane-free autonomous driving and applied  it in a single-agent ring-road environment, using the Deep Deterministic Policy Gradient (DDPG) RL algorithm for policy optimization. 
Our MDP framework is highly influenced by that of~\cite{Karalakou2023} to apply MCTS in our autonomous lane-free driving environment.

\section{Our Approach}
In this section, we first present the lane-free traffic environment and the agent's objectives, then define our MDP for MCTS in which our agent operates, and discuss aspects relevant to NN-Guided MCTS in lane-free traffic.

\subsection{Lane-Free Traffic Environment}
\begin{figure}[t]
    \centering
    \includegraphics[width=0.8\textwidth, keepaspectratio]{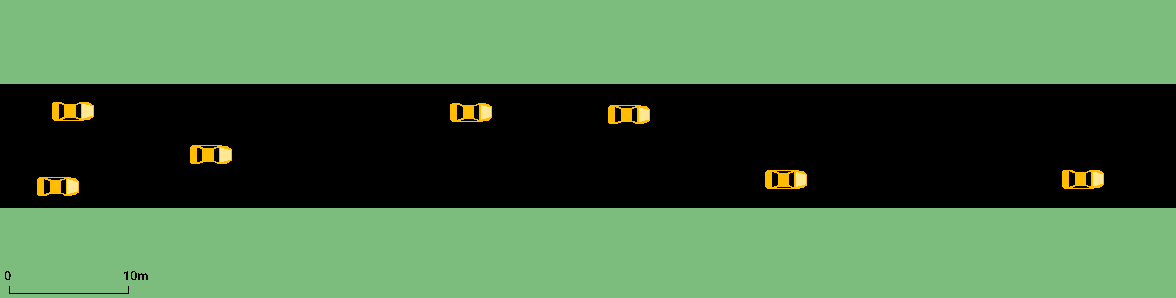} 
    \caption{Snapshot of the simulation environment.}
    \label{fig:sumo}
\end{figure}

In this study, we model an open highway environment to simulate lane-free traffic scenarios with multiple automated vehicles, as shown in Figure~\ref{fig:sumo}.
Each vehicle is an agent that observes its surrounding and has to update its action at every time-step, following an MDP formulation.
Agents (vehicles) execute the MCTS algorithm independently, from their own local view,
tackling the following two objectives, namely: follow a desired speed and avoid collisions. 
Each agent monitors its own longitudinal and lateral positions $(p_x, p_y)$ and speeds $(v_x, v_y)$ as well as the positions and speeds of nearby vehicles. 
All vehicles are standardized in size and movement dynamics, and each one possesses a {\em desired speed} sampled from a predefined range $[v_{d,\min}, v_{d,\max}]$.
This represents the individual speed objective the agent wishes to reach and maintain.
Control inputs for the agent include longitudinal and lateral accelerations $(a_x, a_y)$, influencing gas/brake behaviour and left/right steering respectively.

For this lane-free traffic scenario, we use an extension of the Simulation of Urban MObility (SUMO) tool tailored for lane-free traffic~\cite{Troullinos2022ExtendingSUMO}.
The open highway environment is a straight road where vehicles constantly enter the highway's entry point following a flow demand value, i.e., how many vehicles to be entered per hour, and then exit once they reach the end of the road.

\subsection{State Space}
Firstly, we define the state space $\mathcal{S}$. 
A vehicle $c$ is characterized by attributes:  $c = \{p_x, p_y, v_x, v_y,\\ l, w, v_d\}$, where each attribute is defined as presented in Table~\ref{tab:state}.
\begin{table}[b]
\centering
\begin{tabular}{ll}
    \hline
    {Parameter} & {Definition} \\ \hline
    \(p_x\)  ($m$)& Position in x-axis \\
    \(p_y\)  ($m$)& Position in y-axis \\
    \(v_x\) (\( {m/s} \))& Speed in x-axis \\
    \(v_y\) (\( {m/s} \))& Speed in y-axis \\
    \(l\) \ \ ($m$) & Vehicle length \\
    \(w\) \  ($m$)& Vehicle width \\
    \(v_d\) (\( {m/s} \)) & Desired speed \\ \hline
\end{tabular}
\caption{Vehicle's state information.}
\label{tab:state}
\end{table}

At any moment $t$, the state of a vehicle $c$ includes its own characteristics, along with information about nearby vehicles within a visibility distance $d$. A state $s$ is described as $s = \{c_n, \Gamma = [c_1 \ldots c_m]\}$, where $c_n$ is the ego vehicle, and $\Gamma$ is the set of neighboring vehicles located in front and behind the ego vehicle within a prescribed observational longitudinal distance. 
An instrumental aspect in approaches for lane-free traffic is {\em nudging}, that is, vehicles observe and react to nearby vehicles located on the back as well, instead of typical car-following behaviour where vehicles adjust their behaviour only according to leading vehicles.
The related MDP formulation of~\cite{Karalakou2023} also incorporates such state representations with isotropic information regarding both leading and tailing vehicles.
A terminal state is reached when the ego vehicle collides with another vehicle or exceeds the road boundaries.

\subsection{Action Space}
\begin{figure}[t]
    \centering
    \includegraphics[width=0.8\textwidth, keepaspectratio]{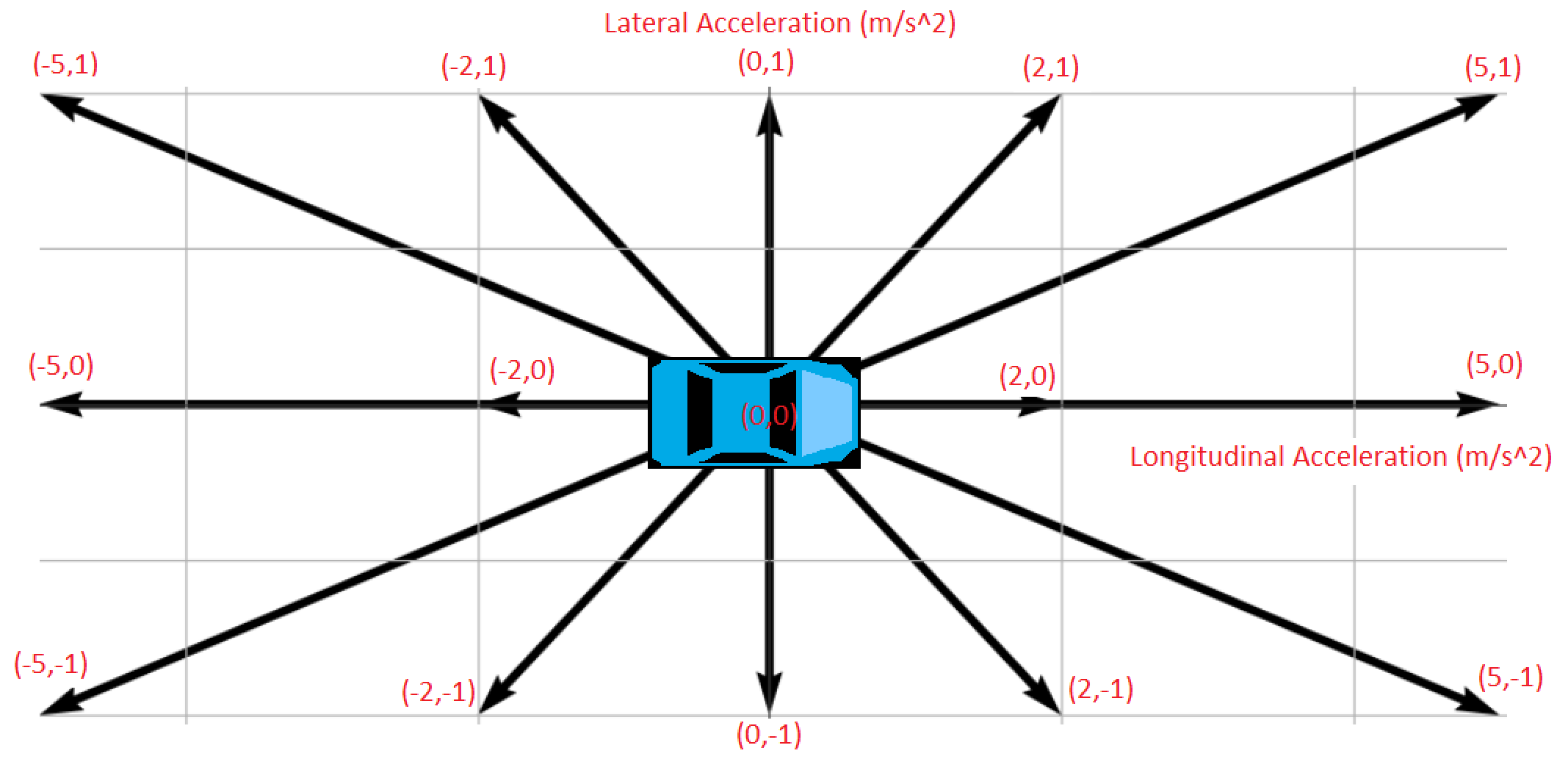} 
    \caption{Action Space consists of $15$ discrete actions in the $2$-dimensional space $\{a_x, a_y\}$.}
    \label{fig:actions}
\end{figure}

For constructing a search tree within our algorithm, the action space $\mathcal{A}$ is discretized into a finite set of possible accelerations. An action $\alpha$ is defined as a pair of selected accelerations, $\alpha = \{a_x, a_y\}$, where $a_x$ and $a_y$ are chosen from predefined sets:
\begin{itemize}
    \item Longitudinal acceleration values ($\text{m/s}^2$): $\alpha_x = \{-5, -2, 0, 2, 5\}$
    \item Lateral acceleration values ($\text{m/s}^2$): $\alpha_y = \{-1, 0, 1\}$
\end{itemize}
This discretization is visualized in Figure~\ref{fig:actions} and results in a set of $15$ distinct actions,
providing a manageable set of maneuvers to choose from.

\subsection{Reward Function}

The reward function guides the algorithm towards finding the optimal policy by quantifying the outcomes of different actions under various states. The primary goals are collision avoidance and maintaining a speed close to the {\em desired} one. 
Therefore, given a state $s$ reached after the rollout, the reward to be backpropagated is calculated as follows.

\subsubsection{Collision Avoidance Objective}

For collision avoidance, we have a collision ``negative reward'' (penalty) term $r_c(s)$.
If a collision occurs during rollout, the simulated state is terminal.
The penalty is lower for collisions occurring later (deeper) in the simulation, as they 
are, intuitively, less likely to occur in practice. That is,
\begin{equation}
    \label{eq:4.1}
    r_c(s) = \left\{
    \begin{array}{ll}
        -\frac{D}{n_s}, & \text{in case of a collision occurrence}\\		
        0, & \text{otherwise}
    \end{array}
    \right.
\end{equation}

\noindent where \( D \) is a positive constant and \( n_s \) is the number of steps after the current state (i.e., the tree depth), as such calibrating the collision's significance based on how far ahead in the simulation it occurs.
If no collision occurs in the rollout, then $r_c(s)$ has no influence.

Additionally, artificial potential fields~\cite{Karalakou2023} measure collision risk based on relative positions and speeds between two vehicles $i$,$j$, calculated by the function $f_{ij}$.
Besides adding a more informed element for proactive collision avoidance, they also promote safety-related gaps that the vehicles should adhere to. 
If $f_{ij}=0$, then vehicles $i$,$j$ already maintain proper distances, while for $f_{ij}<0$, ego vehicle $i$ should react accordingly.
The use of fields---in conjunction with our state representation---additionally enables { nudging} behaviour, meaning that vehicles can react to faster tailing vehicles usually to give them space to overtake.
A more detailed presentation can be found in~\cite{Karalakou2023}.
As such, we also include a cumulative reward term $r_f(s)$ based on the influence of surrounding vehicles in the observed state.
The related term is given by:
\begin{equation}
    \label{eq:4.1_1}
    r_f(s) = \sum_{j=1}^{\Gamma} f_{ij} \cdot (dx_{ij}\leq d_{max})
\end{equation}

\noindent where $\Gamma$ is the set of neighbors in state $s$, then \( dx_{ij} \) is the longitudinal distance between vehicles \( i \) and \( j \), and $d_{max}$ is an influence zone that prunes the effect of vehicles $j$ that are located outside of the designated area.

\subsubsection{Desired Speed Objective}

For this,
we focus on minimizing the distance between the vehicle’s speed $v_x$ at state $s$ and the desired speed $v_d$. The heuristic for the desired speed maintenance objective is defined as:
    \begin{equation}
    \label{eq:4.2}
    r_v(s) = \frac{\epsilon}{\left| v_x - v_d \right| + \epsilon}
    \end{equation}

\noindent where $\epsilon$ is a  positive constant that avoids division with zero.

\subsubsection{Objectives Integration}

The total reward $r(s)$ for a simulation, considering both collision avoidance and speed maintenance, is calculated as a weighted sum:
\begin{equation} 
    \label{eq:4.3}
    r(s) =  \left\{
    \begin{array}{ll}
        \alpha \cdot r_f(s) + \beta \cdot r_c(s), & r_c(s)\neq 0\\		
        \alpha \cdot r_f(s) + c \cdot r_v(s), & \text{otherwise}
    \end{array}
    \right.
    \end{equation}

\noindent where $\alpha,\beta,c$ are a coefficient weighing the importance of the associated terms.
Therefore, we always have the influence from the fields ($r_f(s)$), and select to either contain the collision-related reward ($r_c(s)$) or motivate the agent to pursue its desired speed ($r_v(s)$) if the situation does not result in a collision.

\subsection{Simulation of Future States in Rollout Phase}

The outcome of a vehicle’s action within the simulation is deterministic. 
When a vehicle is in a particular state and takes an action, the
next state is determined by the commonly used kinematic equations of the double integrator model.
For the longitudinal axis ($x$), the equation for the speed update is: $v_{x}' = v_x + a_x T$.
Then, the vehicle’s updated position is calculated by: $p_{x}' = p_x + v_x T + \frac{1}{2} a_x T^2$, where $p_x$ represents the position at current state $s$, $v_x$ denotes the speed, $a_x$ signifies the applied acceleration at that time-step, and $T$ is the constant time-step duration.
Likewise, the lateral axis ($y$) follows the same updates.

Therefore, when an action $a=\{a_x,a_y\}$ is executed, we can simulate the resulting state $s'$ for the ego vehicle's information. 
However, we need to additionally simulate the neighboring vehicles' state information.
To do so, we assume that they perform a ``neutral'' action---that is, zero acceleration on both axes $a=\{0,0\}$.
Thus, their speeds remain unchanged for the simulation purposes, and their positions are updated using the same kinematic equation: \(p_{x}' = p_x + v_x T\). The simulation outcomes, with scores calculated from reward function $r(s)$, provide valuable information for evaluating the simulated states.

\subsection{NN Pre-Training and Calibration}
\begin{figure}
    \centering
    \includegraphics[width=0.6\textwidth]{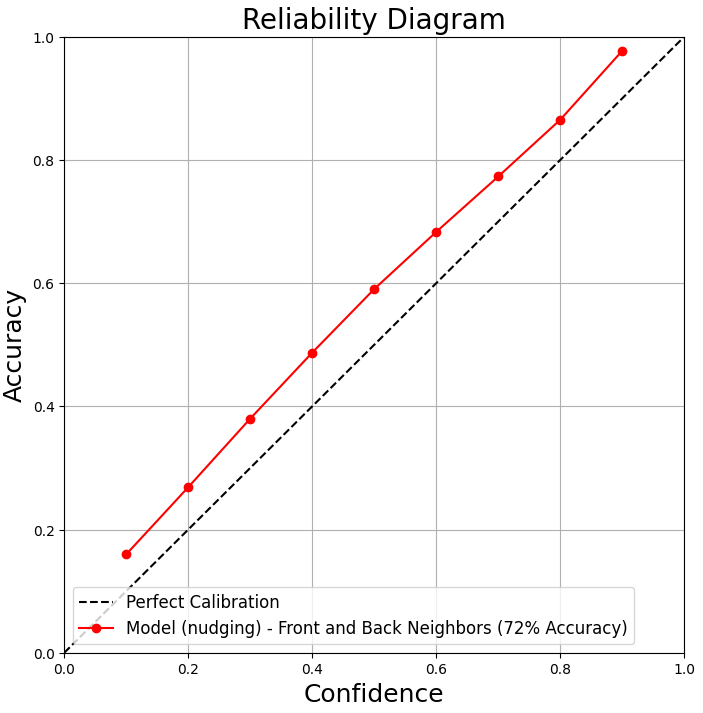} 
    \caption{Reliability Diagram of trained NN's calibration compared to a perfect calibration plotted line.}
    \label{fig:r_d}
\end{figure}

To construct a dataset for learning purposes, we collect data through ``self-play'' simulations using the standard MCTS algorithm. 
This dataset includes inputs (states) and corresponding labels (actions determined by MCTS), where each input element has a vector form of fixed size for compliance with the NN.
After training, if we directly evaluate the calibration on a validation dataset, we obtain a low $72\%$ score for accuracy, even after empirical fine-tuning of the NN architecture and training-related parameters.
However, 
this was to be expected, since we 
(naively)
employed this metric instead of addressing the inherent probabilistic nature of the collected MCTS data.
As such, we instead evaluate the trained NN with the use of tools that better capture the 
accuracy of NNs trained from probabilistic data. 
To this end, we employ {\em Reliability Diagrams}~\cite{GuoPleissSunWeinberger2017}, and partition the dataset into $10$ bins for evaluation based on the probability of the action taken.
As shown in Figure~\ref{fig:r_d}, when we examine the partitioned space, our model closely aligns with the ``ideal'' calibration, indicating that the trained NN 
is quite adept.
More details regarding relevant background, data collection and evaluation process, and other technical information can be found in the
Appendix Section~\ref{app:sec:a}.

\subsection{Technical Aspects of NN integration into MCTS}

As discussed in~\cite{GaoHaywardMuller2017}, a practical limitation of incorporating NN predictions is the computational overhead of the online forward pass for predictions.
Authors briefly mentioned that they resorted to an asynchronous technique to pipeline the procedure.
In this work, while we are more inclined towards the investigation of the acceleration benefits and impact of the NN-guided MCTS in our domain, we have already done some groundwork to improve upon actual execution times.
After early attempts based on {memoization} for $P(s,a)$ on each tree node,
we harness the parallelization capabilities of a GPU through preemptive {\em batch predictions} for child nodes (at the root tree state before the initiation of MCTS) up to a prespecified tree-depth, and therefore have the information for $P(s,a)$ readily available.
More details can be found in the Appendix Sections~\ref{app:sec:b},~\ref{app:sec:c}.

\section{Experimental Evaluation}
This section details our environment setup and simulation parameters. 
We compare the performance of four algorithms: plain MCTS, MCTS enhanced with nudging, NN-Guided MCTS, and standalone NN, across various scenarios to evaluate their effectiveness.

Table~\ref{tab:simul} 
summarizes the parameters of our lane-free simulation environment.
Other parameter settings can be found in Appendix Section~\ref{app:sec:c}.

\begin{table}
    \centering
    \begin{tabular}{lll}
        \hline
        Parameter & Value \\ \hline
        Simulation Time ($sec$)   & $3600$   \\
        Simulation Time-Step ($sec$)  & $0.25$ \\
        Vehicle Length ($m$) & $3.5$  \\
        Vehicle Width ($m$)& $1.6$ \\
        Demand flow (${veh/h}$) & ${5400, 9400, 12000}$ \\ 
        Departure Speed ($m/s$)& $25$ \\
        Desired Speed ($m/s$) & $[25, 35]$ \\
        Road Length ($m$)& $500$ \\ 
        Road Width ($m$) & $10.2$ \\ \hline
    \end{tabular}
    \caption{Simulation Parameters.}
    \label{tab:simul}
\end{table}
    
\subsection{Algorithms, Metrics, Environment and MCTS Settings}

\subsubsection{Examined Algorithms}
Our experiments evaluate several variants. 
{\em MCTS} is a basic implementation where the ego vehicle is aware of only the vehicles in front. 
{\em MCTS (nudging)} enhances the state information about vehicles behind the ego vehicle, improving state awareness and resulting in nudging behaviour that significantly improves upon the resulting policy, as also suggested in other studies~\cite{Papageorgiou2021}.
{\em NN-MCTS} combines NN predictions with {\em MCTS (nudging)} for data-driven guidance. 
Lastly, {\em NN} is the straightforward NN approach without any search methods, used to evaluate its standalone decision-making efficacy.

\subsubsection{Assessed Metrics}
The performance of our algorithms is assessed using two key metrics.
Collisions count the total incidents where two vehicles collide with each other, with fewer collisions indicating better decision-making.
Speed Average calculates the mean speed of all vehicles, ideally close to $30 m/s$ (due to ranging from $25$ to $35 m/s$), reflecting the algorithm's efficiency in maintaining a relatively high speed.

\subsubsection{Examined Flow Demands on Open Highway}
Our algorithms are tested across ranging vehicle flow levels on an open highway to assess their performance under increasingly challenging scenarios. As vehicle flow per hour rises, traffic density increases, complicating decision-making and necessitating more delicate maneuvers to avoid collisions.
Higher traffic density typically reduces average speed, and sometimes can lead to unavoidable collisions for policies that under-perform. 
We evaluate our agents under three flow rates: $5400$ ${veh/h}$ (low traffic), $9400$ ${veh/h}$ (intermediate traffic), and $12000$ ${veh/h}$ (high traffic), each testing the algorithms' ability to maintain efficiency and safety in progressively denser conditions.
\begin{figure}[H]
    \centering
    \includegraphics[width=0.8\textwidth, keepaspectratio]{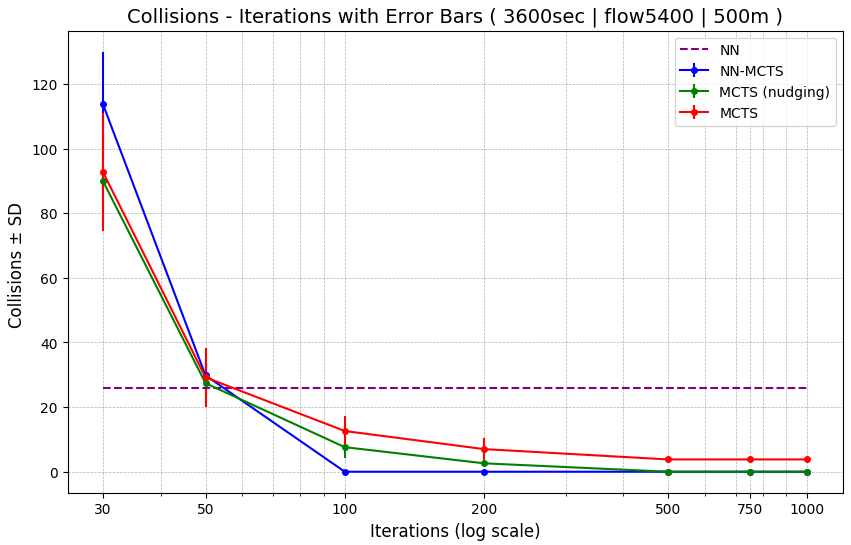} 
    \caption{Graph of Collisions$\pm$SD for NN-MCTS, MCTS (nudging), MCTS and NN across different Iterations for 5400 ${veh/h}$.}
    \label{fig:col1}
\end{figure}

\begin{figure}[H]
\centering
\includegraphics[width=0.8\textwidth, keepaspectratio]{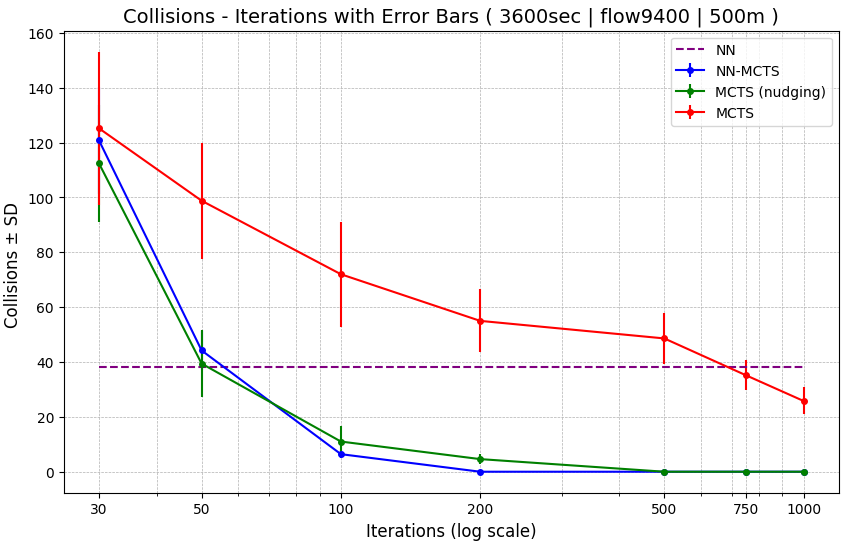} 
\caption{Graph of Collisions$\pm$SD for NN-MCTS, MCTS (nudging), MCTS and NN across different Iterations for 9400 ${veh/h}$.}
\label{fig:col2}
\end{figure}

\begin{figure}[H]
    \centering
    \includegraphics[width=0.8\textwidth, keepaspectratio]{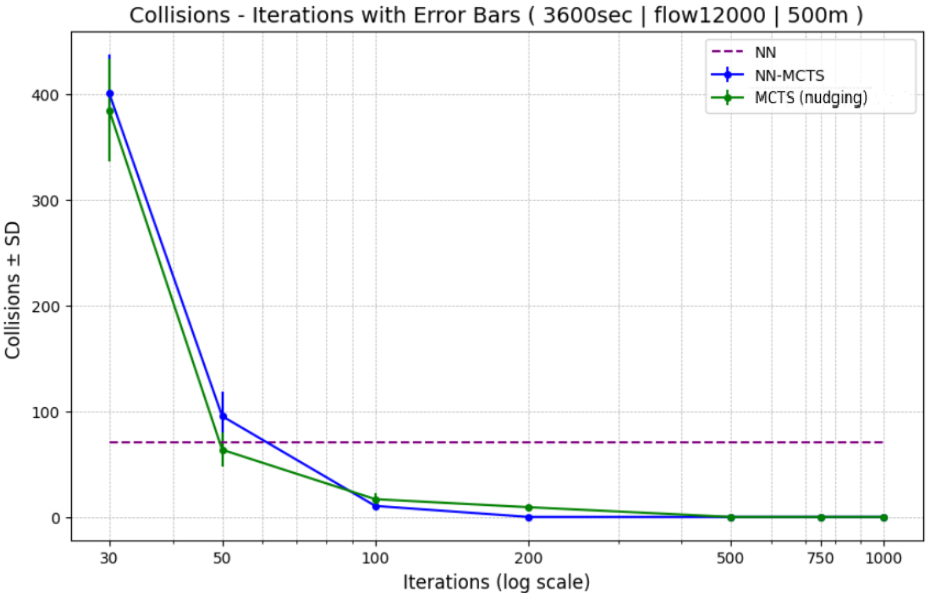} 
    \caption{Graph of Collisions$\pm$SD for NN-MCTS, MCTS (nudging), MCTS and NN across different Iterations for 12000 ${veh/h}$.}
    \label{fig:col3}
\end{figure}

\subsubsection{MCTS Iterations}
In our experimental setup, we vary the number of MCTS iterations for each algorithm, ranging from $30$ to $1000$, to evaluate their performance under different computational constraints.
This approach helps identify the efficiency of each algorithm in finding best actions with minimal iterations.
The goal is to determine the convergence point, where performance peaks and additional iterations yield no notable improvement in result quality. 
By comparing how quickly each algorithm reaches this point and the quality of their results at convergence, we can assess their computational efficiency and efficiency of decision-making.
All results are averaged from $5$ different seeds.

\subsection{Collisions Avoidance Results}
In Figures~\ref{fig:col1}--~\ref{fig:col3}, we present the collisions-iterations diagrams for each demand flow, where accompanying tables detailing the average of collisions and their $\pm$ standard deviations for all algorithms across different iteration samples can be found in the Appendix Section~\ref{app:sec:d}.
We use a logarithmic scale for the iterations axis in our graphs to emphasize performance improvements at lower iteration counts, where significant changes occur, while higher iterations exhibit minimal variation.
From all figures above, we observe that each MCTS algorithm starts with high collision counts, which decrease as iterations increase, indicating more effective decision-making with deeper state exploration.

\subsubsection{Comparison: MCTS vs MCTS (nudging)}
Our comparative analysis shows that MCTS (nudging) consistently outperforms MCTS in reducing collisions across all vehicle flows. MCTS (nudging) achieves stabilization with zero collisions at $500$ iterations for all tested flows, whereas MCTS struggles to reach the best performance, particularly at higher flows.
At $12000$ ${veh/h}$, MCTS is completely incompetent without nudging, resulting in collisions that prevent the simulation from proceeding.
Additionally, MCTS shows larger standard deviations (error bars), indicating less reliable performance. 
These findings demonstrate that MCTS (nudging), just with its enhanced visibility of back neighbors, makes use of the additional information and substantially improves the resulting policy (with nudging effect).

\subsubsection{Comparison: MCTS (nudging) vs NN-MCTS}
NN-MCTS significantly outperforms MCTS (nudging) in reducing collisions more efficiently across all vehicle flows. 
NN-MCTS achieves zero collisions with far fewer iterations: only $100$ iterations at a $5400$ ${veh/h}$ flow, compared to MCTS (nudging), which requires $500$ iterations ($400$ fewer). 
At higher flows of $9400$ and $12000$ ${veh/h}$,
we observe the same phenomenon but naturally requiring more iterations as the problem is more complex.
When NN-MCTS converges first, MCTS (nudging) still experiences collisions, indicating that NN-MCTS's early stabilization marks a phase where plain MCTS is yet to optimize fully and returns inferior results. 
These findings highlight the accelerated efficiency of NN-MCTS due to the impact of leveraging the NN's predictive capabilities.
Notably, at lower iterations ($30$ and $50$), plain MCTS performs slightly better than NN-MCTS.
We attribute this to the initial biases introduced by the NN in NN-MCTS, as the NN is trained on simulations running at $1000$ iterations, meaning that its guidance is optimized for longer search processes. 
The early termination prevents NN-MCTS from fully utilizing its pre-trained knowledge, making it more affected by early cutoffs. 

\begin{figure}[H]
    \centering
    \includegraphics[width=0.8\textwidth, keepaspectratio]{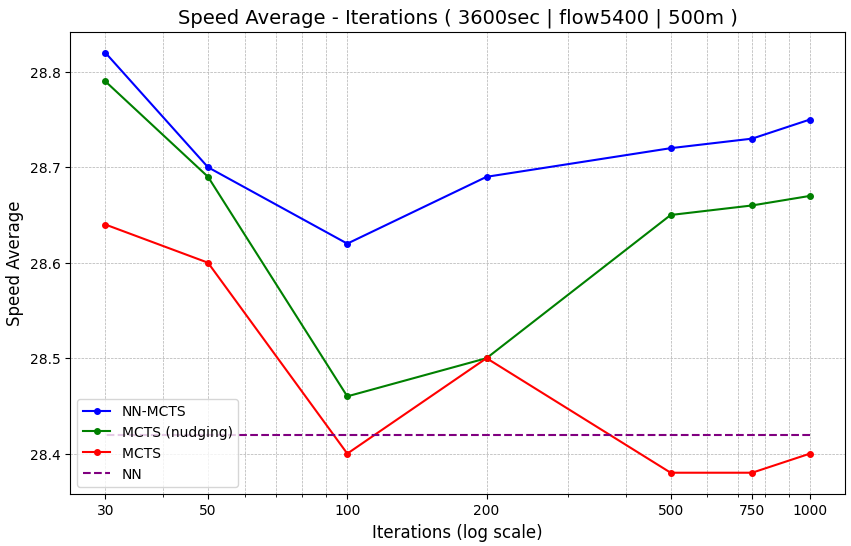} 
    \caption{Graph of Speed Average$\pm$SD for NN-MCTS, MCTS (nudging), MCTS and NN across different Iterations for $5400$ ${veh/h}$.}
    \label{fig:speed1}
\end{figure}

\subsubsection{NN Performance and Comparisons}
The simple NN, which naively predicts the next action, is depicted with a straight-line in graphs 
as it does not involve iterative computations.
It only outperforms other algorithms at low iteration counts 
as these algorithms have not had enough iterations to optimize their solutions. 
However, as the iteration count increases other algorithms delve deeper and  
yield more informed solutions.
Naturally, this is due to the NN's greedy approach, which follows a deterministic strategy without exploring alternative paths.
Despite being trained on data from MCTS (nudging) 
the NN does not replicate expected outcomes due to training inaccuracies and inherent randomness in MCTS data.

\begin{figure}[H]
    \centering
    \includegraphics[width=0.8\textwidth, keepaspectratio]{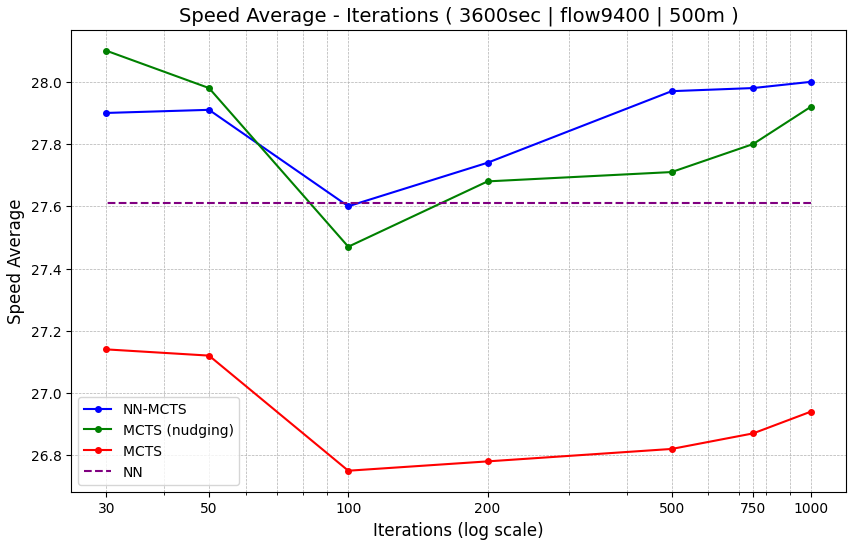} 
    \caption{Graph of Speed Average$\pm$SD for NN-MCTS, MCTS (nudging), MCTS and NN across different Iterations for $9400$ ${veh/h}$.}
    \label{fig:speed2}
\end{figure}

\begin{figure}[H]
    \centering
    \includegraphics[width=0.8\textwidth, keepaspectratio]{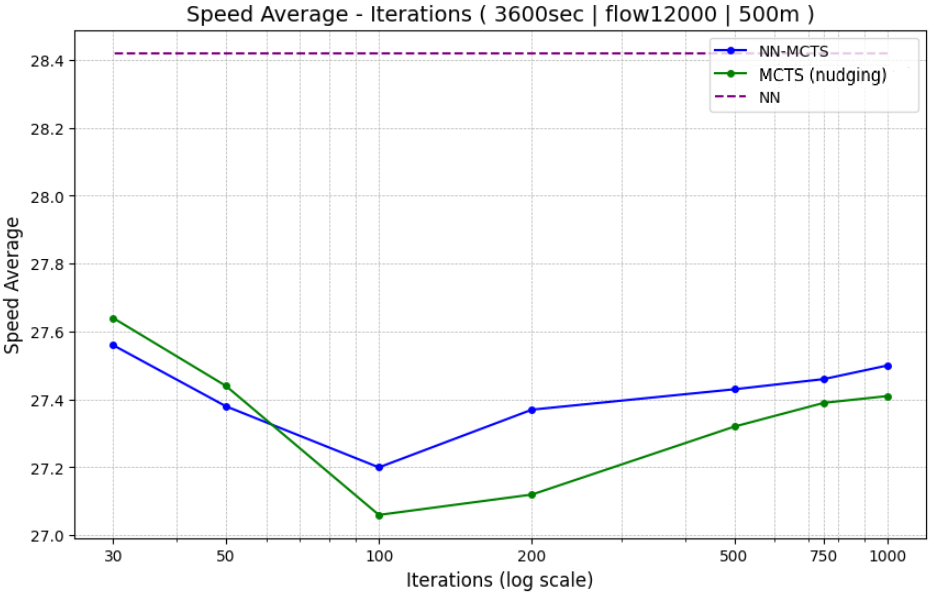} 
    \caption{Graph of Speed Average$\pm$SD for NN-MCTS, MCTS (nudging), MCTS and NN across different Iterations for $12000$ ${veh/h}$.}
    \label{fig:speed3}
\end{figure}

\subsection{Average Speed Results}

Following our discussion, we present the figures for average speed-iterations across all vehicle flows in Figures~\ref{fig:speed1}--~\ref{fig:speed3}.
As vehicle flow increases, MCTS-based algorithms show decreased average speeds. 
This trend highlights the impact of higher traffic density, necessitating more cautious driving behaviors to avoid collisions, which results in slower traffic movement and longer delays.
Complementary results are available in the Appendix Section~\ref{app:sec:d}.

The analysis of speed metric aligns with our previous findings from collision results. 
MCTS (nudging) consistently outperforms MCTS across all vehicle flows, demonstrating better average speeds,
particularly at higher densities where MCTS fails to execute effectively.
NN-MCTS marginally surpasses MCTS (nudging), 
showing slight improvements once stabilized, although these enhancements are not as substantial. 
A related observation is that we do not have convergence for the speed metric, i.e., it is marginally improved even when we reach our imposed limit of $1000$ iterations.
Additionally, in contrast to the collision-related results, the standard deviation (error bars) is consistently close to $0$ for all algorithms.
The plain NN, while consistently worse than both NN-MCTS and MCTS (nudging), still performs better than MCTS in most scenarios, further illustrating the limitations of MCTS without the nudging effect. 
Overall, the findings confirm that the improvements observed in collision metrics also apply to speed, 
reaffirming the superior performance of MCTS (nudging) and NN-MCTS.

At lower iteration counts ($30$ and $50$), a slight improvement in 
speed average is observed across all algorithms. This improvement is due to the algorithms not yet stabilizing their decision-making processes, resulting in more "reckless" decisions that lead to higher speeds.
However, these decisions are not sustainable or safe, as evident by the associated  collisions. The vehicles exhibit behaviors such as lack of proper nudging, and non-compliance with safety rules, contributing to short-term improvements in speed but at the cost of increased risk and collisions.

\begin{figure*}[t]
    \centering
    \includegraphics[width=0.95\linewidth]{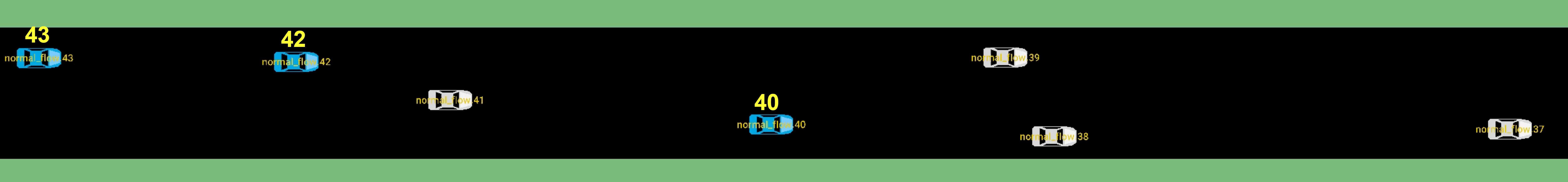}
    \caption{Initial conditions of the example.}
    \label{fig:snap}
\end{figure*}

\subsection{Simulation Execution Example and Vehicle Trajectories}

In this section, we demonstrate the resulting behaviour of the NN-Guided MCTS through the trajectories of the video example found in the supplementary material.
In this example, the vehicles operate with different desired speeds, namely $29$, $32$, and $34 m/s$.
The initial conditions of the scenario are shown in Figure~\ref{fig:snap}, with the blue vehicles being the ones we focus at.
The provided figures provide a clear visualization of the trajectories for position and  speed of the vehicles as they evolve over time in the simulation, with the x axis containing the discrete time-steps.

\begin{figure}[H]
    \centering
    \includegraphics[width=1\textwidth, keepaspectratio]{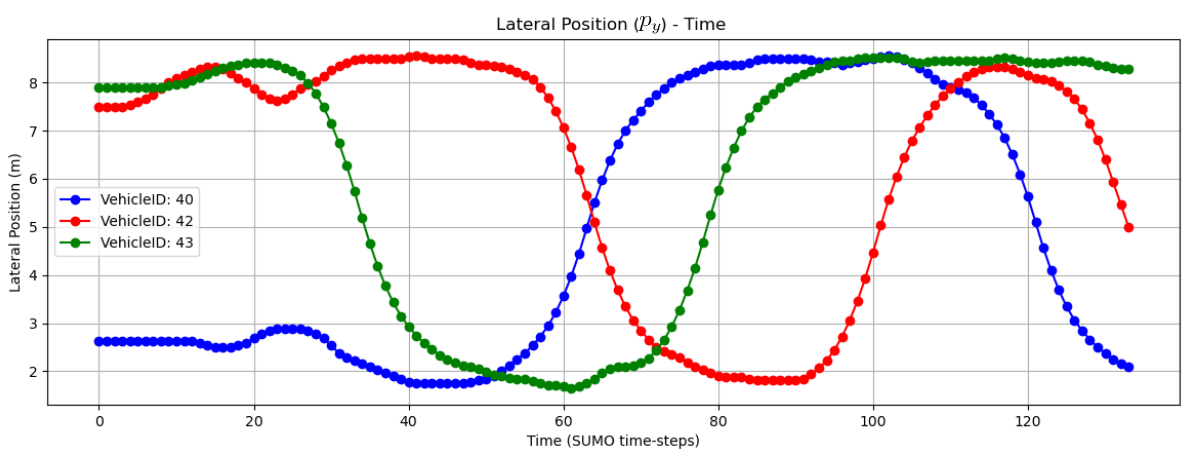} 
    \caption{Lateral position ${p_y}$ ($m$) – Time (time-steps) for vehicles with IDs 40, 42 and 43 from the simulation.}
    \label{fig:section5-figure12}
\end{figure}

\paragraph{Vehicle Trajectories: Lateral Position $\boldsymbol{p_y}$ (m) – Time (time-steps)}
Figure~\ref{fig:section5-figure12} shows the lateral positions of the vehicles, mapping their trajectories across the road's width. Rising curves indicate left turns, falling curves indicate right turns, and flat lines show straight movement.
The figure above aligns with the trajectories observed in the simulation video, showcasing the NN-Guided MCTS algorithm's effectiveness. The vehicles not only accelerate to achieve their desired speeds but also perform complex maneuvers and efficiently navigate between other vehicles, maintaining safe distances to avoid collisions.
    
\begin{figure}[H]
    \centering
    \includegraphics[width=1\textwidth, keepaspectratio]{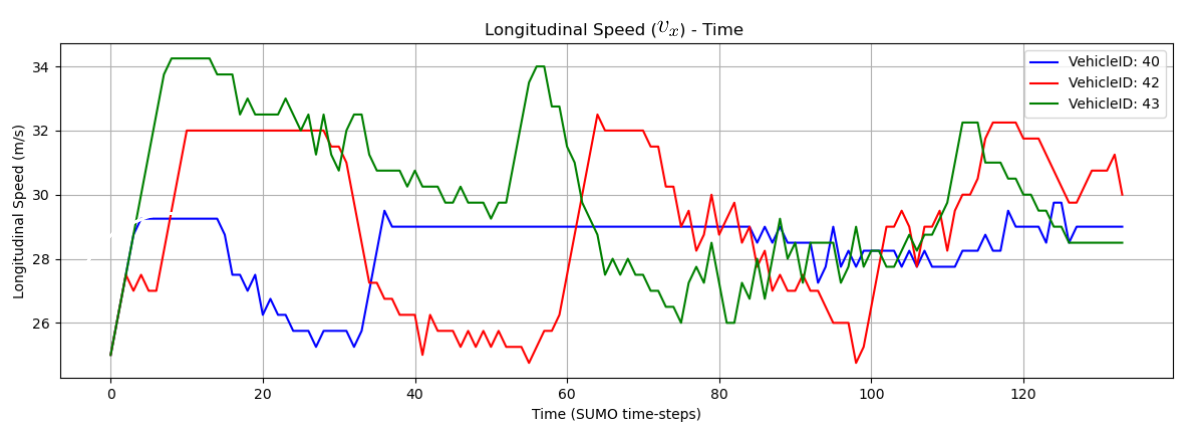} 
    \caption{Longitudinal speed ${v_x}$ ($m/s$) – Time (time-steps) for vehicles with IDs 40, 42 and 43 from the simulation.}
    \label{fig:section5-figure13}
\end{figure}

\paragraph{Vehicle Speed: Longitudinal Speed $\boldsymbol{v_x}$ (m/s) – Time (time-steps)}
Figure~\ref{fig:section5-figure13} depicts the longitudinal speeds. Upward trends signify acceleration, while downward trends indicate deceleration along the x-axis.
The figure above also highlights the variance in vehicle speeds relative to their desired targets. For instance, vehicle 40, with a desired speed of $29 m/s$, maintains lower speeds, while vehicle 43, aiming for $34 m/s$, reaches the highest speeds. This indicates that each vehicle actively tries to match its target speed, with adjustments for braking and maneuvers to avoid collisions.

\begin{figure}[H]
    \centering
    \includegraphics[width=1\textwidth, keepaspectratio]{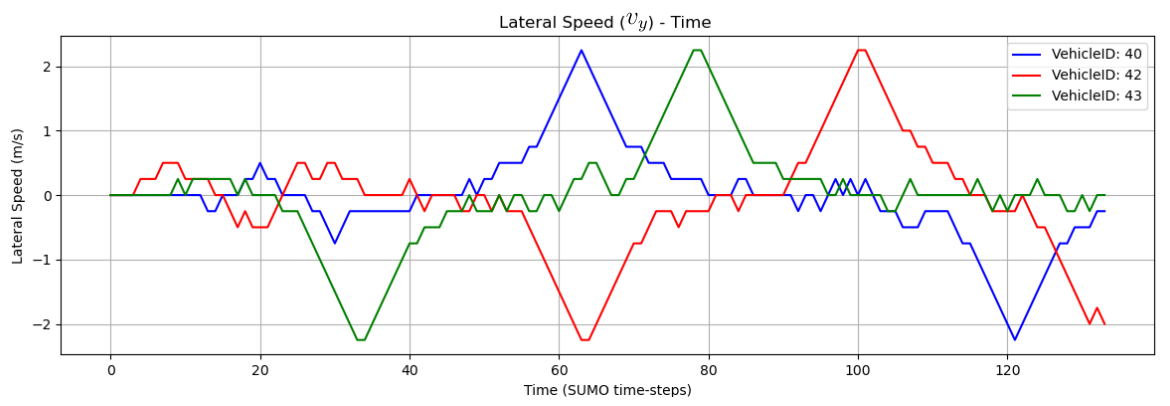} 
    \caption{Lateral Speed ${v_y}$ (m/s) – Time (time-steps) for vehicles with ID's 40, 42 and 43 from the simulation.}
    \label{fig:section5-figure14}
\end{figure}

\paragraph{Lateral speed $\boldsymbol{v_y}$ ($m/s$) – Time (time-steps)}
Figure~\ref{fig:section5-figure14} visualizes lateral speeds, where positive values  indicate leftward movement, and negative values signify rightward movement.
Furthermore, the figure above shows the relationship between lateral speed changes and vehicle positioning. 
Smaller variations result in minor trajectory shifts, whereas larger changes lead to more distinct movements. These observations exhibit the algorithm's ability to adapt to dynamic driving conditions.

\section{Conclusions and Future Work}

In this paper, we developed an MDP for an MCTS algorithm in lane-free traffic environments,
and examined the influence of backward state awareness in nudging behaviour. We further integrated a neural network into the MCTS selection phase, trained through self-play simulations to provide predictive guidance. 
Especially in scenarios with limited iterations, our experimental evaluation shows that integrating NN guidance to MCTS enhanced with nudging, can deliver more sophisticated and better informed results, as plain MCTS may not be able to find the solutions of high quality 
under a designated (realistic) time allowance.

In future work, we can examine neural guidance in other MCTS phases~\cite{Kemmerling2023, deng2022neuralaugmented, song2020dataefficient}, a guided expansion mechanism within MCTS could enable more sophisticated decisions. By selectively expanding only those child nodes identified as promising by the NN and pruning the others, we can efficiently manage increased action space complexity. 
In relation to that, extending NN guidance into the simulation phase of MCTS~\cite{Kemmerling2023, silver2018general} can improve the  predictive accuracy of simulations.
In our domain, this involves simulating neighbor vehicle actions based on the NN’s learned policy, resulting in more accurate and strategic simulations.
Moreover, shifting from offline supervised learning to an online learning paradigm~\cite{Kemmerling2023, chen2022optimizing, rinciog2020sheetmetal} is another potential avenue.
These dynamic, iterative approaches, involve continuous policy improvement via MCTS during the actual tree search process, creating a feedback loop between MCTS and the NN.
Additionally, extensions that address Continuous action spaces for MCTS~\cite{Lee_Jeon_Kim_Kim_2020,Kim_Lee_Lim_Kaelbling_Lozano-Perez_2020,yee2016monte} would fare favourably in our inherently continuous lane-free environment domain.

Finally, NN-MCTS has a time overhead due to the combined use of Python for NN predictions and C++ for the MCTS search procedure through Pybind11, therefore not currently representative of the execution time of proper native implementation that would have
to be employed in a real-time system. 
To this end, technical developments towards a native C++ solution (instead of the current hybrid C++/Python environment), combined with the existing practical choices (e.g., batch prediction) would make future approaches much faster in terms of execution times and therefore better conforming to realistic time constraints.

\section*{Acknowledgements}
The research leading to these results has received funding from the European Research Council under the European Union’s Horizon 2020 Research and Innovation programme/ERC Grant Agreement n.[833915], project TrafficFluid.

\bibliographystyle{abbrv}
\bibliography{lf_mcts}

@article{Browne2012MCTSSurvey,
  title={A Survey of Monte Carlo Tree Search Methods},
  author={Browne, Cameron and Powley, Edward and Whitehouse, Daniel and Lucas, Simon and Cowling, Peter and Rohlfshagen, Philipp and Tavener, Stephen and Perez Liebana, Diego and Samothrakis, Spyridon and Colton, Simon},
  journal={IEEE Transactions on Computational Intelligence and AI in Games},
  volume={4},
  number={1},
  pages={1--43},
  year={2012},
  month={Mar},
  url={10.1109/TCIAIG.2012.2186810}
}

@article{Swiechowski2022MCTSReview,
  title={Monte Carlo Tree Search: a review of recent modifications and applications},
  author={{\'S}wiechowski, Maciej and Godlewski, Konrad and Sawicki, Bartosz and Ma{\'n}dziuk, Jacek},
  journal={Artificial Intelligence Review},
  year={2022},
  month={Jul},
  note={Published online},
  eprint={2103.04931},
  eprinttype={arxiv},
  url={https://link.springer.com/article/10.1007/s10462-022-10228-y}
}

@article{Bellman1957Markovian,
  title={A Markovian decision process},
  author={Bellman, Richard},
  journal={Journal of Mathematics and Mechanics},
  volume={},
  number={},
  pages={679--684},
  year={1957},
  publisher={}
}

@inproceedings{Kocsis2006BanditMCPlanning,
  title={Bandit Based Monte-Carlo Planning},
  author={Kocsis, Levente and Szepesv{\'a}ri, Csaba},
  booktitle={Machine Learning: ECML 2006},
  pages={282--293},
  year={2006},
  publisher={Springer Berlin Heidelberg},
  address={Berlin, Heidelberg},
  isbn={978-3-540-46056-5}
}

@book{GoodfellowBengioCourville2016DeepLearning,
  title={Deep Learning},
  author={Goodfellow, Ian J. and Bengio, Yoshua and Courville, Aaron},
  year={2016},
  publisher={MIT Press},
  address={Cambridge, MA, USA},
  url={http://www.deeplearningbook.org}
}

@InProceedings{GuoPleissSunWeinberger2017,
  title = 	 {On Calibration of Modern Neural Networks},
  author =       {Chuan Guo and Geoff Pleiss and Yu Sun and Kilian Q. Weinberger},
  booktitle = 	 {Proceedings of the 34th International Conference on Machine Learning},
  pages = 	 {1321--1330},
  year = 	 {2017},
  editor = 	 {Precup, Doina and Teh, Yee Whye},
  volume = 	 {70},
  series = 	 {Proceedings of Machine Learning Research},
  month = 	 {06--11 Aug},
  publisher =    {PMLR},
  url = 	 {https://proceedings.mlr.press/v70/guo17a.html}
}

@article{MoonKimShinHwang2020,
  title={Confidence-Aware Learning for Deep Neural Networks},
  author={Moon, Jooyoung and Kim, Jihyo and Shin, Younghak and Hwang, Sangheum},
  journal={arXiv preprint arXiv:2007.01458},
  year={2020},
  url={https://arxiv.org/pdf/2007.01458.pdf}
}

@article{Kavzoglu2009,
  title={Increasing the accuracy of neural network classification using refined training data},
  author={Kavzoglu, Taskin},
  journal={ResearchGate},
  year={2009},
  address={Department of Geodetic and Photogrammetric Engineering, Gebze Institute of Technology, Muallimkoy Campus, P.K.141, 41400 Gebze-Kocaeli, Turkey},
  url={https://www.researchgate.net/publication/223565557_Increasing_the_accuracy_of_neural_network_classification_using_refined_training_data}
}

@article{GaoHaywardMuller2017,
   author={Gao, Chao and Hayward, Ryan and Müller, Martin},

  journal={IEEE Transactions on Games}, 

  title={Move Prediction Using Deep Convolutional Neural Networks in Hex}, 

  year={2018},

  volume={10},

  number={4},

  pages={336-343},


  doi={10.1109/TG.2017.2785042}
}

@article{Kemmerling2023,
  title={Beyond games: a systematic review of neural Monte Carlo tree search applications},
  author={Kemmerling, Marco and L{\"u}tticke, Daniel and Schmitt, Robert H},
  journal={Applied Intelligence},
  volume={54},
  number={1},
  pages={1020--1046},
  year={2024},
  publisher={Springer}
}

@article{Karafyllis2021,
  title={Two-dimensional cruise control of autonomous vehicles on lane-free roads},
  author={Karafyllis, I. and Theodosis, D. and Papageorgiou, M.},
  journal={arXiv preprint arXiv:2103.12205},
  year={2021},
  month={Mar},
  archivePrefix={"arXiv"},
  primaryClass={math.OC},
  url={https://arxiv.org/abs/2103.12205}
}

@inproceedings{Troullinos2021,
  title={Collaborative multiagent decision making for lane-free autonomous driving},
  author={Troullinos, D. and Chalkiadakis, G. and Papamichail, I. and Papageorgiou, M.},
  booktitle={AAMAS '21},
  year={2021},
  pages={1335--1343},
  organization={International Foundation for Autonomous Agents and Multiagent Systems},
  address={Virtual Event, United Kingdom},
  
}

@ARTICLE{Yanumula2022,
  author={Yanumula, Venkata Karteek and Typaldos, Panagiotis and Troullinos, Dimitrios and Malekzadeh, Milad and Papamichail, Ioannis and Papageorgiou, Markos},
  journal={IEEE Transactions on Intelligent Vehicles}, 
  title={Optimal Trajectory Planning for Connected and Automated Vehicles in Lane-Free Traffic With Vehicle Nudging}, 
  year={2023},
  volume={8},
  number={3},
  pages={2385-2399},
  }

@article{Karalakou2023,
  title={Deep reinforcement learning reward function design for autonomous driving in lane-free traffic},
  author={Karalakou, A. and Troullinos, D. and Chalkiadakis, G. and Papageorgiou, M.},
  journal={Systems},
  volume={11},
  number={3},
  year={2023},
  
}

@article{Papageorgiou2021,
  author={Papageorgiou, Markos and Mountakis, Kyriakos-Simon and Karafyllis, Iasson and Papamichail, Ioannis and Wang, Yibing},
  journal={Proceedings of the IEEE}, 
  title={Lane-Free Artificial-Fluid Concept for Vehicular Traffic}, 
  year={2021},
  volume={109},
  number={2},
  pages={114-121},
}

@inproceedings{Ha2020,
  title={Vehicle Control with Prediction Model Based Monte-Carlo Tree Search},
  author={Ha, Timothy and Cho, Kyunghoon and Cha, Geonho and Lee, Kyungjae and Oh, Songhwai},
  booktitle={17th International Conference on Ubiquitous Robots (UR)},
  year={2020},
  pages={303--308},
  url={https://doi.org/10.1109/UR49135.2020.9144958}
}

@article{Chen2020,
  title={Driving Maneuvers Prediction Based Autonomous Driving Control by Deep Monte Carlo Tree Search},
  author={Chen, Jienan and Zhang, Cong and Luo, Jinting and Xie, Junfei and Wan, Yan},
  journal={IEEE Transactions on Vehicular Technology},
  volume={69},
  number={7},
  pages={7146--7158},
  year={2020},
  url={https://doi.org/10.1109/TVT.2020.2991584}
}

@inproceedings{Lei2021,
  title={KB-Tree: Learnable and Continuous Monte-Carlo Tree Search for Autonomous Driving Planning},
  author={Lei, Lanxin and Luo, Ruiming and Zheng, Renjie and Wang, Jingke and Zhang, JianWei and Qiu, Cong and Ma, Liulong and Jin, Liyang and Zhang, Ping and Chen, Junbo},
  booktitle={2021 IEEE/RSJ International Conference on Intelligent Robots and Systems (IROS)},
  address={Prague, Czech Republic},
  pages={4493--4500},
  year={2021},
  publisher={IEEE Press},
  doi={10.1109/IROS51168.2021.9636442},
  url={https://doi.org/10.1109/IROS51168.2021.9636442}
}

@inproceedings{Volpi2017,
  title={Towards event-based MCTS for autonomous cars},
  author={Catenacci Volpi, Nicola and Wu, Yan and Ognibene, Dimitri},
  booktitle={2017 Asia-Pacific Signal and Information Processing Association Annual Summit and Conference (APSIPA ASC)},
  year={2017},
  pages={420--427},
  organization={IEEE}
}

@inproceedings{Troullinos2022ExtendingSUMO,
  title={Extending sumo for lane-free microscopic simulation of connected and automated vehicles},
  author={Troullinos, D. and Chalkiadakis, G. and Manolis, D. and Papamichail, I. and Papageorgiou, M.},
  booktitle={SUMO Conference Proceedings},
  volume={3},
  pages={95--103},
  year={2022},
  month={Sep},
  url={https://doi.org/10.52825/scp.v3i.110}
}

@article{RodriguezCardiff2022PythonOpenFOAM,
  title={A general approach for running Python codes in OpenFOAM using an embedded pybind11 Python interpreter},
  author={Rodriguez, Simon and Cardiff, Philip},
  journal={arXiv preprint arXiv:2203.16394},
  year={2022},
  note={Preprint submitted to OpenFOAM Journal},
  eprint={2203.16394},
  archivePrefix={arXiv},
  primaryClass={cs.CE},
  comment={Preprint submitted to OpenFOAM Journal},
  subjects={Computational Engineering, Finance, and Science (cs.CE)},
  url={https://arxiv.org/abs/2203.16394}
}

@misc{tensorflow2015whitepaper,
title={ {TensorFlow}: Large-Scale Machine Learning on Heterogeneous Systems},
url={https://www.tensorflow.org/},
note={Software available from tensorflow.org},
author={
    Mart\'{i}n~Abadi and
    Ashish~Agarwal and
    Paul~Barham and
    Eugene~Brevdo and
    Zhifeng~Chen and
    Craig~Citro and
    Greg~S.~Corrado and
    Andy~Davis and
    Jeffrey~Dean and
    Matthieu~Devin and
    Sanjay~Ghemawat and
    Ian~Goodfellow and
    Andrew~Harp and
    Geoffrey~Irving and
    Michael~Isard and
    Yangqing Jia and
    Rafal~Jozefowicz and
    Lukasz~Kaiser and
    Manjunath~Kudlur and
    Josh~Levenberg and
    Dandelion~Man\'{e} and
    Rajat~Monga and
    Sherry~Moore and
    Derek~Murray and
    Chris~Olah and
    Mike~Schuster and
    Jonathon~Shlens and
    Benoit~Steiner and
    Ilya~Sutskever and
    Kunal~Talwar and
    Paul~Tucker and
    Vincent~Vanhoucke and
    Vijay~Vasudevan and
    Fernanda~Vi\'{e}gas and
    Oriol~Vinyals and
    Pete~Warden and
    Martin~Wattenberg and
    Martin~Wicke and
    Yuan~Yu and
    Xiaoqiang~Zheng},
  year={2015},
}

@misc{chollet2015keras,
  title={Keras},
  author={Chollet, Fran\c{c}ois and others},
  year={2015},
  howpublished={\url{https://keras.io}},
}

@misc{chetlur2014cudnn,
      title={cuDNN: Efficient Primitives for Deep Learning}, 
      author={Sharan Chetlur and Cliff Woolley and Philippe Vandermersch and Jonathan Cohen and John Tran and Bryan Catanzaro and Evan Shelhamer},
      year={2014},
      eprint={1410.0759},
      archivePrefix={arXiv},
      primaryClass={cs.NE},
      url={https://arxiv.org/abs/1410.0759}, 
}

@article{ghorpade2012gpgpu,
  title={GPGPU Processing in CUDA Architecture},
  author={Ghorpade, Jayshree and Parande, Jitendra and Kulkarni, Madhura and Bawaskar, Amit},
  journal={Advanced Computing: An International Journal},
  volume={3},
  number={1},
  pages={105},
  year={2012},
  month={Jan},
  doi={10.5121/acij.2012.3109},
  email={jayshree.aj@gmail.com, jitendra.parande@sungard.com, madhurak25@gmail.com, amitbawaskar@gmail.com},
  affiliation={Department of Computer Engineering, MIT College of Engineering, Pune University, India; SunGard Global Technologies, India},
  url={https://arxiv.org/ftp/arxiv/papers/1202/1202.4347.pdf}
}

@article{chen2022optimizing,
  author    = {Chen, Y.-Q. and Chen, Y. and Lee, C.-K. and Zhang, S. and Hsieh, C.-Y.},
  title     = {Optimizing Quantum Annealing Schedules with Monte Carlo Tree Search Enhanced with Neural Networks},
  journal   = {Nature Machine Intelligence},
  volume    = {4},
  number    = {3},
  year      = {2022},
  pages     = {269--278},
  publisher = {Nature Publishing Group},
  url       = {https://arxiv.org/abs/2004.02836},
  doi       = {Insert-DOI-if-available}
}

@article{deng2022neuralaugmented,
  author    = {Deng, H. and Yuan, X. and Tian, Y. and Hu, J.},
  title     = {Neural-augmented Two-stage Monte Carlo Tree Search with Over-sampling for Protein Folding in HP Model},
  journal   = {IEEJ Transactions on Electrical and Electronic Engineering},
  volume    = {17},
  number    = {5},
  year      = {2022},
  pages     = {685--694},
  publisher = {Wiley Online Library},
  url       = {https://onlinelibrary.wiley.com/doi/abs/10.1002/tee.23556},
  doi       = {10.1002/tee.23556}
}

@inproceedings{rinciog2020sheetmetal,
  author    = {Rinciog, A. and Mieth, C. and Scheikl, P. M. and Meyer, A.},
  title     = {Sheet-metal Production Scheduling Using AlphaGo Zero},
  booktitle = {Proceedings of the Conference on Production Systems and Logistics: CPSL 2020},
  year      = {2020},
  url       = {https://www.repo.uni-hannover.de/handle/123456789/9732},
  organization = {Leibniz Universität Hannover},
  address   = {Hannover, Germany}
}

@article{silver2018general,
  author    = {Silver, D. and Hubert, T. and Schrittwieser, J. and Antonoglou, I. and Lai, M. and Guez, A. and Lanctot, M. and Sifre, L. and Kumaran, D. and Graepel, T. and Lillicrap, T. and Simonyan, K. and Hassabis, D.},
  title     = {A General Reinforcement Learning Algorithm That Masters Chess, Shogi, and Go Through Self-Play},
  journal   = {Science},
  volume    = {362},
  number    = {6419},
  year      = {2018},
  pages     = {1140--1144},
  publisher = {American Association for the Advancement of Science},
  url       = {https://pubmed.ncbi.nlm.nih.gov/30523106/},
  doi       = {10.1126/science.aar6404}
}

@article{song2020dataefficient,
  author    = {Song, S. and Chen, H. and Sun, H. and Liu, M.},
  title     = {Data Efficient Reinforcement Learning for Integrated Lateral Planning and Control in Automated Parking System},
  journal   = {Sensors},
  volume    = {20},
  number    = {24},
  year      = {2020},
  url       = {https://www.mdpi.com/1424-8220/20/24/7297},
  doi       = {10.3390/s20247297}
}

@article{Lee_Jeon_Kim_Kim_2020, title={Monte-Carlo Tree Search in Continuous Action Spaces with Value Gradients}, volume={34}, url={https://ojs.aaai.org/index.php/AAAI/article/view/5885}, DOI={10.1609/aaai.v34i04.5885}, abstractNote={&lt;p&gt;Monte-Carlo Tree Search (MCTS) is the state-of-the-art online planning algorithm for large problems with discrete action spaces. However, many real-world problems involve continuous action spaces, where MCTS is not as effective as in discrete action spaces. This is mainly due to common practices such as coarse discretization of the entire action space and failure to exploit local smoothness. In this paper, we introduce Value-Gradient UCT (VG-UCT), which combines traditional MCTS with gradient-based optimization of action particles. VG-UCT simultaneously performs a global search via UCT with respect to the finitely sampled set of actions and performs a local improvement via action value gradients. In the experiments, we demonstrate that our approach outperforms existing MCTS methods and other strong baseline algorithms for continuous action spaces.&lt;/p&gt;}, number={04}, journal={Proceedings of the AAAI Conference on Artificial Intelligence}, author={Lee, Jongmin and Jeon, Wonseok and Kim, Geon-Hyeong and Kim, Kee-Eung}, year={2020}, month={Apr.}, pages={4561-4568} }

@article{Kim_Lee_Lim_Kaelbling_Lozano-Perez_2020, title={Monte Carlo Tree Search in Continuous Spaces Using Voronoi Optimistic Optimization with Regret Bounds}, volume={34}, url={https://ojs.aaai.org/index.php/AAAI/article/view/6546}, DOI={10.1609/aaai.v34i06.6546}, abstractNote={&lt;p&gt;Many important applications, including robotics, data-center management, and process control, require planning action sequences in domains with continuous state and action spaces and discontinuous objective functions. Monte Carlo tree search (MCTS) is an effective strategy for planning in discrete action spaces. We provide a novel MCTS algorithm (&lt;span style=&quot;font-variant: small-caps;&quot;&gt;voot&lt;/span&gt;) for deterministic environments with continuous action spaces, which, in turn, is based on a novel black-box function-optimization algorithm (&lt;span style=&quot;font-variant: small-caps;&quot;&gt;voo&lt;/span&gt;) to efficiently sample actions. The &lt;span style=&quot;font-variant: small-caps;&quot;&gt;voo&lt;/span&gt; algorithm uses Voronoi partitioning to guide sampling, and is particularly efficient in high-dimensional spaces. The &lt;span style=&quot;font-variant: small-caps;&quot;&gt;voot&lt;/span&gt; algorithm has an instance of &lt;span style=&quot;font-variant: small-caps;&quot;&gt;voo&lt;/span&gt; at each node in the tree. We provide regret bounds for both algorithms and demonstrate their empirical effectiveness in several high-dimensional problems including two difficult robotics planning problems.&lt;/p&gt;}, number={06}, journal={Proceedings of the AAAI Conference on Artificial Intelligence}, author={Kim, Beomjoon and Lee, Kyungjae and Lim, Sungbin and Kaelbling, Leslie and Lozano-Perez, Tomas}, year={2020}, month={Apr.}, pages={9916-9924} }

@inproceedings{yee2016monte,
  title={Monte Carlo Tree Search in Continuous Action Spaces with Execution Uncertainty.},
  author={Yee, Timothy and Lis{\`y}, Viliam and Bowling, Michael H and Kambhampati, S},
  booktitle={IJCAI},
  pages={690--697},
  year={2016}
}

\newpage
\appendix
\section*{Appendix}

\section{Probabilistic Learning and Calibration in Neural Networks}\label{app:sec:a}
In a supervised learning problem, accuracy~\cite{Kavzoglu2009} measures the percentage of correct predictions, while confidence~\cite{MoonKimShinHwang2020} indicates the probability that a given prediction is correct, providing insight into the reliability of individual predictions.

Learning from probabilistic data involves handling situations where the same input can lead to different outcomes with certain probabilities.
In stochastic processes and probabilistic scenarios (like MCTS), traditional accuracy metrics can be misleading, as they may unfairly penalize the model for making reasonable predictions that do not align perfectly with the true labels due to inherent data uncertainty.
This necessitates the use of alternative metrics to evaluate model performance more accurately.

Model calibration~\cite{GuoPleissSunWeinberger2017} is used to evaluate a neural network (NN)’s output probabilities, ensuring they accurately reflect true outcome probabilities.
This evaluation ensures that the predicted confidence level $\hat{P}$ aligns with the actual likelihood of correctness. Ideal calibration is expressed mathematically as $P(\hat{Y} = Y | \hat{P} = p) = p$ for all $p$ in the range $[0, 1]$. Achieving it is challenging, especially in a multiclass setting, due to the continuous nature of the probability variable.

Reliability diagrams~\cite{GuoPleissSunWeinberger2017} visualize model calibration by plotting predicted confidence against actual accuracy.
Predictions are divided into $M$ equally sized bins based on confidence scores, representing the range $\left(\frac{m-1}{M}, \frac{m}{M}\right]$.
Actual accuracy for each bin $B_m$ is calculated as: $\text{acc}(B_m) = \frac{1}{|B_m|} \sum_{i \in B_m} 1(\hat{y}_i = y_i)$, where  $\hat{y}_i$ and $y_i$ are the predicted and true labels.
Average confidence for each bin is: $\text{conf}(B_m) = \frac{1}{|B_m|} \sum_{i \in B_m} \hat{p}_i$, where $\hat{p}_i$ is the confidence level for each prediction.

Perfect calibration occurs when accuracy equals average confidence in each bin.
Confidence histograms~\cite{GuoPleissSunWeinberger2017} complement reliability diagrams by showing the distribution and number of samples across confidence bins, providing a visual summary of prediction confidences across the dataset.

\subsection{Data Collection and Pre-training}
Continuing with the integration process of the NN to MCTS, we must first gather our dataset for training the network.
To construct the NN for learning purposes, we collect data through self-play simulations using the plain MCTS (nudging) algorithm at 1000 iterations for high quality solutions. This dataset includes inputs (states) and corresponding labels (best actions determined by MCTS). Each input state consists of specific attributes:

\textbf{Ego vehicle values:} Characterized by parameters \(\{p_{\text{y}}, u_x, u_y, w,l, d_u\}\) (6 values); excluding the $x$ position ($p_x$), as it does not influence the decision-making process.

\textbf{Neighbor vehicle values:} Account for the $4$ nearest vehicles in front and the 4 behind within a 50-meter visibility range, defined by parameters \(\{dx, dy, u_x, u_y, w, l, d_u\}\) for each vehicle, (total 56 values). The values $dx$ and $dy$ are the distances in positions longitudinal and lateral relative to the ego vehicle.

If there are more than $4$ vehicles either in front or behind, the closest ones are prioritized. If fewer than 4 vehicles are present, ``virtual'' neighbors with predefined parameters ;so the NN can distinguish them from real neighbors; are introduced, ensuring the state vector's completeness. The final dataset consists of $63$ elements: $62$ values represent a state, and $1$ value denotes the best action.

\subsection{Training and Accuracy}
After generating our dataset, we created and trained our NN classifier to predict the most likely action given a state from the SUMO simulation. Despite optimization and fine-tuning, the NN's accuracy only reached $72\%$.
This raised concerns about the network’s learning capability.

The inherent probabilistic nature of MCTS data, due to random actions rollouts in simulation phase, resulted in identical states leading to different best actions.
The NN, however, simplifies this by predicting the most probable action, producing a deterministic policy.
This discrepancy explains the low accuracy. To ensure our NN is functioning correctly and learning effectively from the data, we  validate its performance with the use of reliability diagrams 
that reflect the probabilistic nature of the learned policy.

\subsection{Reliability Diagrams for Neural Network Evaluation}
To evaluate our model’s calibration, we divided the dataset into 10 bins based on the probability of the most probable prediction for each sample, ranging from $[0, 0.1)$ up to $[0.9, 1]$.
We calculated the actual confidence and accuracy for each bin, aiming for the ideal where confidence equals accuracy.

\subsubsection{Reliability Diagram}
We used reliability diagrams to visualize our model’s calibration against the perfect calibration line. As shown in figure below, our model closely aligns with the ideal calibration line, indicating it is well-calibrated.

\begin{figure}[H]
    \centering
    \includegraphics[width=0.60\textwidth, keepaspectratio]{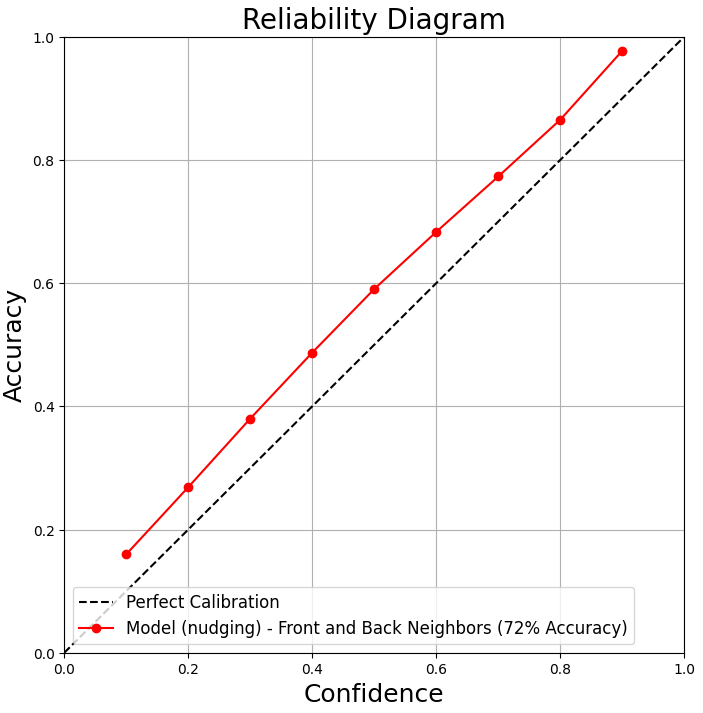} 
    \caption{Reliability Diagram - Model's Calibration and Perfect Calibration.}
    \label{fig:section4-figure4}
\end{figure}
\subsubsection{Confidence Histogram}
Additionally, we created a histogram of confidence to show the distribution of sample predictions across the bins. From figure below, we understand that the samples are distributed correctly, starting from zero and gradually increasing in each next bin. This indicates fewer predictions with low confidence and a proper distribution of prediction confidence levels.
    \begin{figure}[H]
    \centering
    \includegraphics[width=0.9\textwidth, keepaspectratio]{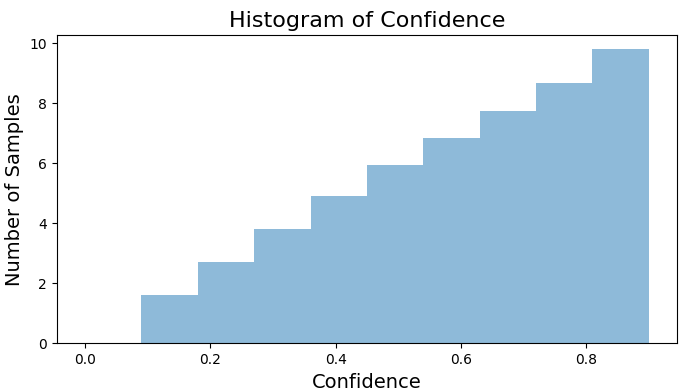} 
    \caption{Histogram of Confidence - Number of Samples per Bin.}
    \label{fig:section4-figure5}
\end{figure}
Finally, another evidence of our network's learning capability and performance is observed when we eliminate the inherent randomness of the MCTS algorithm. By setting the maximum rollout state depth to $1$ (from its current value of $6$), making the next state always deterministic, our NN trained with this data achieved an accuracy of $95.8\%$. This significantly higher accuracy strongly supports our assertion.

\section{Time overhead of NN Predictions}\label{app:sec:b}
In this section we detail how our NN produces its predictions \( P(s, a) \) and the gradual process of refining the functionality and speed of our algorithm. 
\subsection{Single Predictions in Current State}
Initially, our process involved generating predictions for each state in MCTS run-time. The predictive function took the current state and returned a probability distribution of 15 possible actions. This output was then used in the MCTS’s selection phase to determine the best action.

\subsection{Inefficiencies and Speed Improvements}
The initial approach was slow because it repeatedly calculated predictions from start for each selection phase, leading to many redundant calculations. To address this, we stored predictions after the first visit to each node (memoization). 
When revisiting a state, the algorithm retrieved the stored predictions, significantly reducing redundant computations and speeding up the process.

Further optimization focused on transitioning from single to batch predictions. Initially, each prediction call involved overhead from repeatedly invoking a Python script from MCTS run-time which has been written in C++. By adopting batch processing, we utilized the NN’s capability to process multiple inputs simultaneously. This change minimized script call overhead and maximized computational efficiency, resulting in a faster, more streamlined algorithm.

\subsection{Batch Predictions in Child Expansion}
Our strategy for implementing batch predictions involved generating predictions in batches of 15, corresponding to each potential child action of a node. When expanding a node, we computed the predictions for its 15 potential child states and stored each prediction within the respective child nodes. This approach ensured that when revisiting any of these child states, we could directly use the previously calculated predictions, eliminating the need for recalculations. An exception to this method was the root node, which required a single prediction to determine its initial set of predictions.

In terms of speed improvement, shifting to batch processing for 15 predictions at once resulted in significant efficiency gains. The average time required for a single prediction was 13 times faster compared to making 15 individual predictions.

\section{Algorithm Implementation}\label{app:sec:c}

\subsection{Navigation in Autonomous Driving Decision Space}
To construct the MCTS tree for our autonomous driving environment, nodes represent specific states $s$. Starting from the current state as the root node, each node is connected by edges that represent actions, defined as pairs of longitudinal and lateral accelerations. From any given state, there are 15 possible actions, resulting in 15 child nodes. When a new action is taken, it transitions from the current state to a new one. During the simulation phase, the value for each state is calculated using the reward function $r(s)$. This value is then backpropagated through the tree, updating the average reward of each state accordingly.

\subsection{NN Predictions}
\subsubsection{Batch Predictions in Algorithm’s Start}
To enhance the speed of our predictions, we transitioned from generating small batches to calculating predictions for the entire MCTS state space in one go, storing them in an external predictions matrix. This method maximizes the NN batch processing capabilities and eliminates prediction calculations during run-time. Initially, we construct the complete tree state space up to a predefined maximum depth, typically depth 3, based on experimental observations that deeper states are rarely visited for expansion, thus there is no need for calculating their predictions.

\subsubsection{Optimizing Predictions Storage and Retrieval}
By generating predictions for the entire tree in one batch and storing them in a matrix, we significantly reduce the time needed for predictions. When the algorithm runs, it retrieves the necessary predictions from this matrix, just by one memory access, bypassing the need for repeated calculations. This approach ensures that we make the script call only once before the MCTS algorithm starts running, leveraging the efficiency of batch processing. Although the batch processing method involves pre-calculating predictions for states that might not be visited, thus they are redundant, the overall performance enhancement makes this trade-off acceptable.
In Algorithm~\ref{agl2}, we provide a high level implementation of the NN-Guided MCTS algorithm and the selection function it employs (Algorithm~\ref{func11}).
\begin{algorithm2e}[h]
    \SetAlgoLined
    \SetNlSty{textbf}{}{}
    \SetAlgoNlRelativeSize{0}
    \DontPrintSemicolon
    \caption{A High Level Pseudocode of NN-Guided MCTS }
    \label{agl2}
    \KwIn{${current\ state}$}
    {$model \gets \text{load trained neural network once }()$\;
        
    $iterations \gets 0$\;
    
    $root \gets \text{initialize tree }(current\ state)$\; 
    
    {$predictions\ \gets \text{ batch predictions }(model,data\ tree)$\;}
    
    \While{$iterations \leq \textsc{max iterations}$}{
        {
            \If{$state\ depth \leq \textsc{max depth}$}{
                
                $state\ prediction \gets predictions\ [state\ index]$\;
                
                $selected\ state \gets \text{select}(root,\ state\ prediction)$\;
            }
            \Else{
                
                {$state\ prediction\gets \text{ single prediction }(model,data\ single)$\;}
                
                $selected\ state \gets \text{select }(root,\ state\ prediction)$\;
            }
        }
        
        \If{$\text{is terminal } (selected\ state)$}{ 
            
            \textbf{continue}\; 
        }
        $expanded\ state \gets \text{expand }(selected\ state)$\; 
        
        $score \gets \text{simulate }(expanded\ state)$\; 
        
        $\text{backpropagate }(expanded\ state, score)$\; 
        
        $iterations \gets iterations + 1$\;	
    }

            \KwRet{$best\ action$}
    }
\end{algorithm2e}

\begin{algorithm2e}[h]
    \SetAlgoLined
    \SetNlSty{textbf}{}{}
    \SetAlgoNlRelativeSize{0}
    \DontPrintSemicolon
    \caption{Selection Phase of NN-Guided MCTS}
    \label{func11}
    \KwIn{$state,\ {prediction}$}
    $children \gets \text{get children }(state)$\;
    \For{$child \in children$}{
        {$score \gets \text{PUCT }(state,\ child,\ prediction)\ $\;}
    }
    $best child \gets argmax_{child}\ (score) $\;
    
    \KwRet{$best\ child$}
\end{algorithm2e}

\subsubsection{Speed Improvement}
This refined approach led to significant speed improvements. The batch prediction method is 10.2 times faster than the previous 15-batch approach and 132.6 times faster than the original single prediction method. By pre-calculating predictions and storing them in a matrix, we avoid runtime calculations, streamlining the entire process and achieving substantial performance gains. 
In Table~\ref{tab:pred} we can see detailed time metrics for each batch of predictions.
\begin{table}
    \centering
    \begin{tabular}{lll}
        \hline
        Predictions Batch Size & Tree Depth & Time needed in (ms) \\ \hline
        1  & 1 & 26 \\
        16  & 2 & 30\\
        241 & 3 &44 \\
        3616 & 4 &182 \\ \hline
    \end{tabular}
    \caption{Time in milliseconds needed for varying Predictions Batches.}
    \label{tab:pred}
\end{table}

\subsubsection{Decision on Tree Depth}
We observed experimentally that our state space tree, after passing depth 3 tends to explore states mostly in width and not so much in depth. This means, that almost always it reached only up to depth 4 and very rarely depth 5.  This remained the case not only when operating with few iterations, but also when it was tested with high number of iterations, thus more time to explore in depth if it had been more beneficial. So we concluded that we always need the predictions for states in depth 3, so we can expand to depth 4, but we do not need predictions for depth 4 to go to 5. Therefore, we empirically determined that creating our predictions tree up to depth 3 was the most favourable, as further depths resulted in exponentially increased states and computational time without proportional benefits. For example, depth 3 has 241 states, while depth 4 jumps to 3616 states. Hence, depth 3 balances computational efficiency and prediction accuracy. During runtime, if the current state's depth exceeds this predefined depth, a single prediction is made for that state, ensuring that all necessary predictions are available.

\subsection{Deep Neural Network without Search}
In this section, we consider a simple alternative that uses the NN in a greedy manner without any underlying search mechanism like MCTS. At each state, the algorithm makes a single prediction and selects the action with the highest probability. This straightforward approach, while potentially less effective due to its lack of strategic depth, serves as a useful benchmark. By comparing its performance against traditional MCTS and NN-guided MCTS, we can better understand the benefits and limitations of relying solely on NN knowledge for decision-making.

\subsection{Agent and Hyper-parameter Setup}

In this section, we outline the key parameters and technical elements relevant to our agent and the MCTS algorithm. 
The development of the NN was facilitated by Tensorflow-Keras~\cite{tensorflow2015whitepaper,chollet2015keras} for batch processing acceleration using CUDA~\cite{ghorpade2012gpgpu} and cudnn~\cite{chetlur2014cudnn}. 
Integration into the MCTS framework which was implemented in C++, was achieved through Pybind11~\cite{RodriguezCardiff2022PythonOpenFOAM}, allowing Python scripts to run during C++ runtime. Tables~\ref{tab:nn},~\ref{tab:mcts},~\ref{tab:packages} below provide detailed information on the NN architecture, MCTS algorithm parameters and training parameters respectively.

\begin{table}[h]
    \centering
    \begin{tabular}{ll}
        \hline
        Parameter & Value \\ \hline
        Input Layer & 62 (one MCTS state)\\
        Hidden Layers & 512, (0.3 dropout), 256, (0.3 dropout), 128 \\
        Output Layer & 15 (MCTS possible actions)\\
        Type & Feed Forward\\
        Activation Function & ReLU (hidden layers), Softmax (output layer)\\
        Optimizer & Adam\\
        Batch Size& 64\\
        Epochs & 50\\
        Learning Rate & start 0.001 and descending\\
        Accuracy & 72\%(rollout depth 6) , 95.8\% (rollout depth 1)\\
        Front Neighbors (dataset) & 4\\
        Back Neighbors (dataset) & 4\\
        Visibility Target (m) & 50 \\ \hline
    \end{tabular}
    \caption{NN Architecture and Training Parameters.}
    \label{tab:nn}
\end{table}

\begin{table}[h]
    \centering
    \begin{minipage}{0.45\linewidth} 
        \centering
        \begin{tabular}{ll}
            \hline
            Parameter & Value \\ \hline
            D  & 10\\
            \(\epsilon\)  & 1\\
            Min Visits  & 5\\
            Max Rollout Depth& 6\\
            Lateral Actions & {-1, 0, 1}\\
            Longitudinal Actions & {-5, -2, 0 , 2, 5}\\ \hline
        \end{tabular}
        \caption{MCTS Parameters.}
        \label{tab:mcts}
    \end{minipage}
    \hfill 
    \begin{minipage}{0.45\linewidth} 
        \centering
        \begin{tabular}{ll}
            \hline
            Parameter & Value \\ \hline
            CPU & Ryzen 5 3600\\
            RAM & 16 GB\\
            GPU & RTX 3060 4GB \\

            OS & Windows 10\\
            cudnn capability & 8.6\\
            Python Version & 3.10.8\\
            Tensorflow Version & 2.9.1\\ \hline
        \end{tabular}
        \caption{Technical specifications and versions for utilized packages.}
        \label{tab:packages}
    \end{minipage}
\end{table}

\section{Additional Results}\label{app:sec:d}

In Tables~\ref{tab:col1},~\ref{tab:col2},~\ref{tab:col3}, we provide the exact numerical results regarding collisions associated with the figures in the main paper.
Then,
in Table~\ref{tab:nn_col}, we include the collision results for the NN-only variant.

\begin{table}[H]
    \centering
    \begin{tabular}{cccc}
        \hline
        NN-MCTS & MCTS (nudging) & MCTS & Iterations \\ \hline
        0 & 0 & $3.8\pm0.84$ & 1000 \\
        0 & 0 & $3.8\pm0.84$ & 750 \\
        0 & $\mathbf{0}$ & $\mathbf{3.8\pm0.84}$ & $\mathbf{500}$ \\
        0 & ${2.6\pm0.9}$ & $7\pm3.39$ & 200 \\
        $\mathbf{0}$ & $7.6\pm3.4$ & $12.6\pm4.77$ & $\mathbf{100}$ \\
        ${29.8\pm3.9}$ & $27.4\pm6.3$ & $29.2\pm9.04$ & 50 \\
        $113.8\pm16.2$ & $90\pm14.6$ & $92.8\pm18.28$ & 30 \\ \hline
    \end{tabular}
    \caption{Comparative Results of Collisions$\pm$SD for NN-MCTS, MCTS (nudging), and MCTS across different Iterations for 5400 ${veh/h}$. }
    \label{tab:col1}
\end{table}

\begin{table}[H]
\centering
\begin{tabular}{cccc}
    \hline
    NN-MCTS & MCTS (nudging) & MCTS & Iterations \\ \hline
    0 & 0 & $\mathbf{25.8\pm4.92}$ & $\mathbf{1000}$ \\
    0 & 0 & $35.2\pm5.54$ & 750 \\
    0 & $\mathbf{0}$ & $48.6\pm9.37$ & $\mathbf{500}$ \\
    $\mathbf{0}$ & ${4.6\pm1.82}$ & $55\pm11.47$ & $\mathbf{200}$ \\
    ${6.4\pm1.52}$ & $11\pm5.52$ & $72\pm19.07$ & 100 \\
    $44.2\pm7.09$ & $39.4\pm12.3$ & $98.8\pm21.1$ & 50 \\
    $120.8\pm19.43$ & $112.4\pm21.33$ & $125.2\pm27.93$ & 30 \\ \hline
\end{tabular}
\caption{Comparative Results of Collisions$\pm$SD for NN-MCTS, MCTS (nudging), MCTS across different Iterations for $9400$ ${veh/h}$. }
\label{tab:col2}
\end{table}

\begin{table}[H]
    \centering
    \begin{tabular}{cccc}
        \hline
        NN-MCTS & MCTS (nudging) & MCTS & Iterations \\ \hline
        0 & 0 & - & 1000 \\
        0 & 0 & - & 750 \\
        0 & $\mathbf{0}$ & - & $\mathbf{500}$ \\
        $\mathbf{0}$ & ${9.2\pm2.58}$ & - & $\mathbf{200}$ \\
        ${10.4\pm2.6}$ & $16.8\pm5.4$ & - & 100 \\
        $95.4\pm23.1$ & $63.6\pm16.21$ & - & 50 \\
        $400.8\pm37.29$ & $384.8\pm48.33$ & - & 30 \\ \hline
    \end{tabular}
    \caption{Comparative Results of Collisions$\pm$SD for NN-Guided MCTS, MCTS (nudging), and MCTS across different Iterations for $12000$ ${veh/h}$. }
    \label{tab:col3}
\end{table}
\begin{table}[H]
    \centering
    \begin{tabular}{cccc}
        \hline
        5400 veh.flow/h. & 9400 veh.flow/h. & 12000 veh.flow/h.\\ \hline
        22 & 35  & 61   \\ \hline
    \end{tabular}
    \caption{NN Collisions for each flow demand ${veh/h}$. }
    \label{tab:nn_col}
\end{table}

With regards to speed, the numerical results can be found in Table~\ref{tab:speed}.
There, the calculated standard deviation (SD) values were consistently close to $0$ and therefore we omitted their report.

\begin{figure}[H]
    \centering
    \includegraphics[width=0.8\textwidth, keepaspectratio]{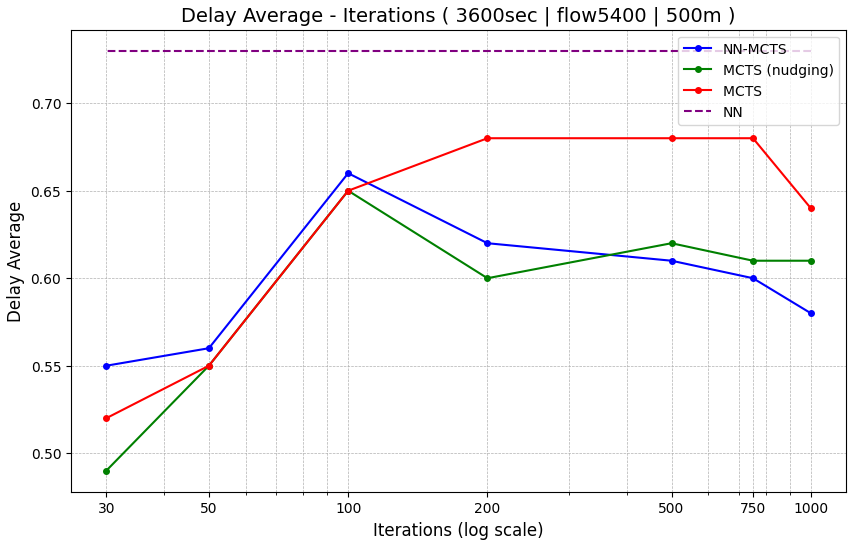} 
    \caption{Delay Average$\pm$SD for NN-MCTS, MCTS (nudging), MCTS and NN across different iterations for 5400 ${veh/h}$.}
    \label{fig:del1}
\end{figure}

\begin{figure}[H]
    \centering
    \includegraphics[width=0.8\textwidth, keepaspectratio]{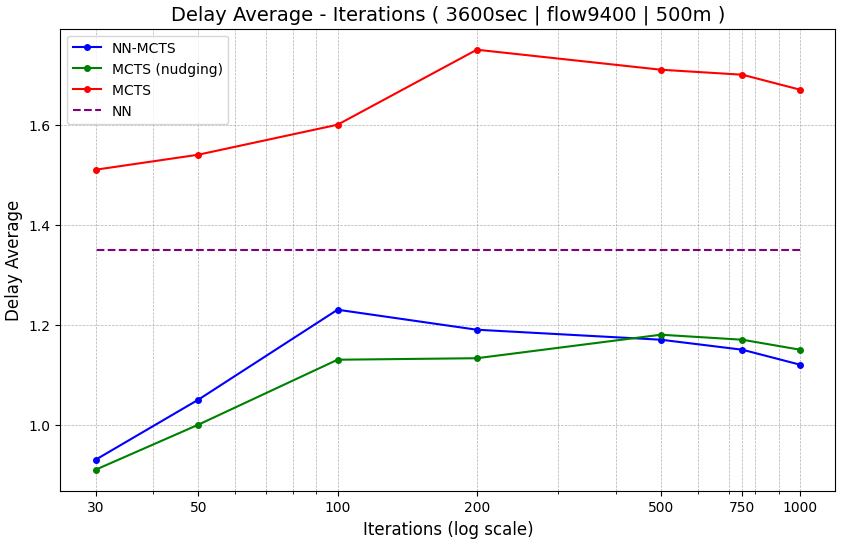} 
    \caption{Delay Average$\pm$SD for NN-MCTS, MCTS (nudging), MCTS and NN across different iterations for $9400$ ${veh/h}$.}
    \label{fig:del2}
\end{figure}

\begin{figure}[H]
    \centering
    \includegraphics[width=0.8\textwidth, keepaspectratio]{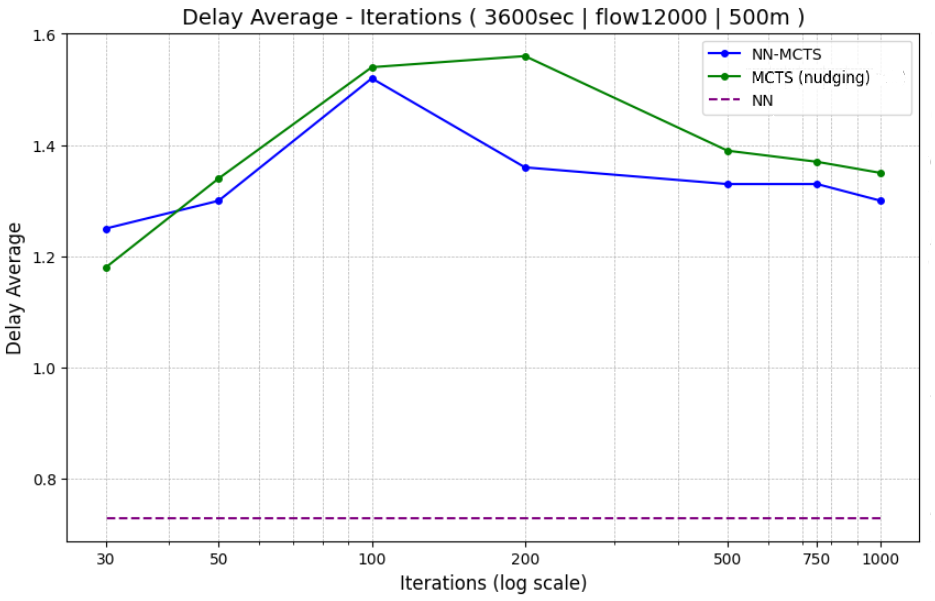} 
    \caption{Delay Average$\pm$SD for NN-MCTS, MCTS (nudging), MCTS and NN across different iterations for $12000$ ${veh/h}$.}
    \label{fig:del3}
\end{figure}

An additional metric to evaluate our methods is delay, which  measures the difference between actual travel time and ideal travel time based on desired speeds, with lower values indicating better policies, where more vehicles better follow their desired speed objective.
In Figures~\ref{fig:del1}--~\ref{fig:del3} and Table~\ref{tab:del}, we report on these results across all the methods.
There, we have consistent findings, again that isotropic state information improves upon delay times (due to nudging), and the NN positively impacts the solution quality especially after a certain number of iteration.
As in the speed related results, we observed very small SD values across all variants.

\begin{table}[H]
\centering
    \begin{tabular}{ccccc}
        \hline
        ${veh/h}$ & NN-Guided MCTS & MCTS (nudging) & MCTS & NN \\ \hline
        5400 & 28.75 & 28.67 & 28.4 & 28.42 \\
        9400 & 28 & 27.92 & 26.94 & 27.61 \\
        12000 & 27.5 & 27.41 & - & 26.98 \\ \hline
    \end{tabular}
    \caption{Speed Average in m/s for NN-Guided MCTS, MCTS (nudging), MCTS and NN for different Vehicle Flows.}
    \label{tab:speed}
\end{table}


\begin{table}[H]
    \centering
    \begin{tabular}{ccccc}
        \hline
        ${veh/h}$ & NN-Guided MCTS & MCTS (nudging) & MCTS & NN \\ \hline
        5400 & 0.58 & 0.61 & 0.64 & 0.73 \\
        9400 & 1.12 & 1.15 & 1.67 & 1.35 \\
        12000 & 1.3 & 1.35 & - & 1.69 \\ \hline
    \end{tabular}
    \caption{Delay Average in ms for NN-Guided MCTS, MCTS (nudging), MCTS and NN for different vehicle flows.}
    \label{tab:del}
\end{table}

\end{document}